\pdfoutput=1

\documentclass[11pt]{article}

\usepackage{acl}
\usepackage{mainlp} %

\usepackage{times}
\usepackage{latexsym}
\usepackage{fontawesome5} %
\usepackage{tikz-dependency}

\usepackage{booktabs}
\usepackage{graphicx}
\usepackage{tabularx}

\usepackage{adjustbox}
\usepackage{arydshln}

\usepackage{subcaption}
\usepackage{xspace}
\usepackage[utf8]{inputenc}

\usepackage{pifont}
\usepackage{makecell}

\newcommand*{\gloss}[1]{\textcolor{gray}{\textit{#1}}}
\newcommand*{\translation}[1]{\textcolor{gray}{`#1'}}
\newcommand{\perturbation}[6]{%
& \texttt{#1} & #3 & #5\\
& #2 & \gloss{#4} & #6 \\[3pt]%
}
\newcommand{\perturbationoverlong}[7]{%
& \texttt{#1} & #3 & #5\\
& #2 & \gloss{#4} & #6 \\
& & & #7 \\[3pt]%
}

\definecolor{purple}{RGB}{128, 13, 166}
\definecolor{green}{RGB}{113, 153, 18}

\newcolumntype{C}{>{\centering\arraybackslash}X}
\newcommand{\mbert}{mBERT\xspace}
\newcommand{\xlmr}{XLM-R\xspace}
\newcommand{\rembert}{RemBERT\xspace}
\newcommand{\mdeberta}{mDeBERTa\xspace}
\newcommand{\mdistilbert}{DistilmBERT\xspace}
\newcommand{\mminilm}{mMiniLM\xspace}

\newcommand{\xsid}{xSID\xspace}
\newcommand{\atis}{MultiATIS++\xspace}
\newcommand{\massive}{MASSIVE\xspace}
\newcommand{\mtop}{MTOP\xspace}

\newcommand{\cmark}{$\checkmark$}

\usepackage{stackengine}
\newcommand{\lmu}{\faMountain}
\newcommand{\mcml}{\faRobot}
\newcommand{\itu}{\faCompass}

\title{Exploring the Robustness of Task-oriented Dialogue Systems for\\ Colloquial German Varieties}
\author{Ekaterina Artemova \textsuperscript{\scriptsize\lmu} \thanks{\ \ Now at Toloka.AI}
\And
Verena Blaschke\kern1pt\stackon[1pt]{\scriptsize\mcml}{\scriptsize\lmu} \\
\textsuperscript{\lmu} MaiNLP, Center for Information and Language Processing, LMU Munich, Germany \\
  \textsuperscript{\mcml} Munich Center for Machine Learning (MCML), Munich, Germany \\
  \textsuperscript{\itu} Department of Computer Science, IT University of Copenhagen, Denmark  \\
{\tt \{verena.blaschke, b.plank\}@lmu.de}
\And
Barbara Plank\kern1pt\stackon[1pt]{\scriptsize\mcml\kern1pt\itu}{\scriptsize\lmu} 
}

\begin{document}
\maketitle
\begin{abstract}

Mainstream cross-lingual task-oriented dialogue (ToD) systems leverage the transfer learning paradigm by training a joint model for intent recognition and slot-filling in English and applying it, zero-shot, to other languages.
We address a gap in prior research, which often overlooked the transfer to lower-resource colloquial varieties due to limited test data.
Inspired by prior work on English varieties, we craft and manually evaluate perturbation rules that transform German sentences into colloquial forms and use them to synthesize test sets in four ToD datasets.
Our perturbation rules cover 18 distinct language phenomena, enabling us to explore the impact of each perturbation on slot and intent performance.
Using these new datasets, we conduct an experimental evaluation across six different transformers.
Here, we demonstrate that when applied to colloquial varieties, ToD systems maintain their intent recognition performance, losing 6\% (4.62 percentage points) in accuracy on average. 
However, they exhibit a significant drop in slot detection, with a decrease of 31\% (21 percentage points) in slot F$_1$ score.
Our findings are further supported by a transfer experiment from Standard American English to synthetic Urban African American Vernacular English.

\end{abstract}

\section{Introduction} \label{sec: intro}

\begin{figure}
    \centering
    \includegraphics[width = 0.95\linewidth]{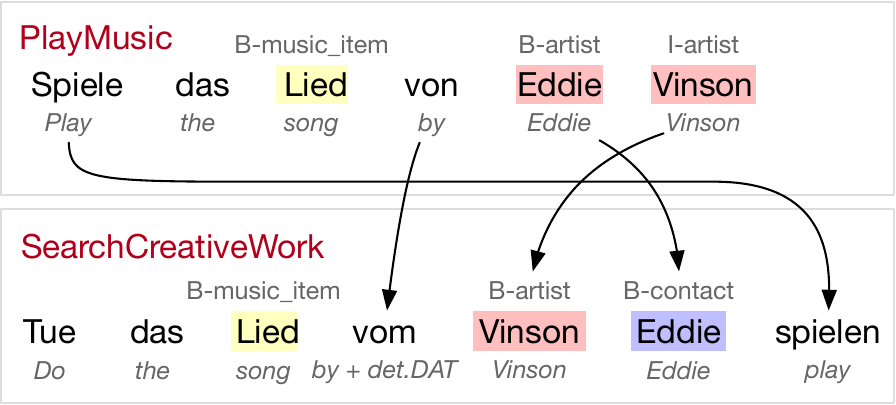}
    \caption{An illustrative example selected from \xsid. The top part displays the intact sentence with gold labels, the bottom part shows the prediction for the perturbed sentence. The perturbations  \texttt{tun\_imperative}, \texttt{article\_name}, \texttt{name\_order}  are applied. There are errors in  predicting the intent and one of the two slots. }
    \label{fig:example}
\end{figure}

The usability of dialog systems heavily relies on the ability to handle user inputs in multiple languages.  
Recent language models (LMs) have become state-of-the-art tools to carry out the primary task-oriented dialogue (ToD) problems, including intent recognition and slot filling. 
What is more, LMs leverage multilingual pre-training to facilitate transfer across languages. 
To achieve this, the mainstream approach involves fine-tuning the LM on a pivot language, commonly English, and subsequently employing the LM in a zero-shot manner to process target languages \cite{pmlr-v119-hu20b}. 

While this language transferring approach has achieved impressive results for many language pairs, its effectiveness is limited when it comes to processing low-resource language varieties and dialects \cite{hedderich-etal-2021-survey}. 
These varieties are often underrepresented in the LM's pre-training data and may not align well with the characteristics of the chosen pivot language. 
Our current understanding of how well modern LMs handle dialects and the extent of disparity between standard languages and dialects remains limited.
Therefore, it is important to assess the performance gap in the first place, as highlighted by  \citet{kantharuban-etal-2023-quantifying} to identify key directions for further development.

Processing (non-standardized) dialects brings unique challenges: large volumes of writing such as newspapers or fiction are rarely produced, and access to conversational data in social media is limited. 
Besides, dialects lack unified spelling rules \cite{millour-fort-2019-unsupervised} and exhibit a high degree of variation over space and time \cite{dunn-wong-2022-stability}.
Finally, dialects may additionally show a significant rate of code-mixing compared to standard languages \cite{muysken2000bilingual}. 

To tackle these challenges, recent studies have introduced techniques that mimic dialectal morphosyntactic variation  through rule-based translation systems which perturb sentences into respective dialect variants in German \cite{gerlach-etal-2022-producing} and English \cite{ziems-etal-2022-value}. 
This approach is highly practical as it avoids the expense of annotating new data while still effectively stress-testing applications like question answering and machine translation \cite{ziems-etal-2023-multi}.
Building on this, we choose ToD as a task where we expect a high level of linguistic variation in real-life application settings \cite{trong2019enabling,aepli-etal-2023-findings}.
We experimentally evaluate how well ToD systems handle dialectal data by simulating dialectal and colloquial variations in English and German to explore the following research questions (RQs). 
\begin{inparaenum}[\bf RQ~1:]
\item How does the LM performance in intent recognition and slot filling change when applied to synthetic dialectal data in both English and German?
\item Considering that each perturbation isolates a specific dialectal phenomenon, which perturbations have the most significant effect?
\item How do LMs differ in terms of robustness to dialectal perturbations? %
\end{inparaenum}
\autoref{fig:example} illustrates our approach.

To address these RQs, we contribute in the following ways: \begin{inparaenum}[\itshape (i)]
  \item We define and implement a set of hand-crafted perturbation rules for translating from Standard German to its spoken varieties (\S \ref{subsec: german_perturb}).
  \item We systematically test a range of perturbations, each representing distinct dialectal phenomena, in two languages, to quantify their individual effect on ToD performance (\S \ref{sec:results}, RQ~1\&2).  
  \item We provide an extensive analysis of joint intent recognition and slot filling experiments using a diverse set of cross-lingual encoders in two languages (\S \ref{sec:results}, RQ~3).  
\end{inparaenum}

We release the code for the perturbation rules and the results of our experimental evaluation for further uptake: \href{https://github.com/mainlp/dialect-ToD-robustness}{\texttt{github.com/mainlp/dialect-\allowbreak{}ToD-\allowbreak{}robustness}}.

\section{Related Work} \label{sec:related}

\paragraph{Robustness of ToD systems.}  
The evaluation of Task-oriented Dialogue (ToD) systems' robustness aims to investigate the generalization capabilities of LMs and their ability to adapt to domain shifts, with a specific focus on English \cite{chang-etal-2021-robustness}.
The robustness of ToD systems has been widely investigated using adversarial attacks, which involve manipulating the gradients and weights of LMs to alter their predictions \cite{cheng-etal-2019-evaluating}.

Nevertheless, white-box methods lack linguistic awareness, making them not easily interpretable \cite{zeng-etal-2021-openattack}.
In contrast, recent black-box methods have emerged that aim to mimic language variation and real-life noise, including speech artifacts and typos, with the primary objective of crafting instances that deceive LMs  \cite{lee2022sgd,liu-etal-2021-robustness,peng-etal-2021-raddle,cho-etal-2022-know}.  

A related line of research focuses on developing defenses against adversarial attacks and enhancing the robustness of ToD systems  by employing techniques such as data augmentations and incorporating regularization terms in the loss function \cite{einolghozati2019improving,sengupta-etal-2021-robustness}.

\paragraph{NLP for dialects and non-standard varieties.}
Previous efforts in processing dialects and non-standard varieties have primarily focused on differentiating between dialects and closely related languages. 
Notably, the VarDial initiative \cite{gaman-etal-2020-report, chakravarthi-etal-2021-findings-vardial, aepli-etal-2022-findings, aepli-etal-2023-findings} has conducted a series of evaluation campaigns aimed at dialect identification and discrimination between similar languages.
Additional research directions in the field include part-of-speech (POS) tagging \cite{hollenstein-aepli-2014-compilation,zampieri-etal-2019-report}, syntactic parsing \cite{blodgett-etal-2018-twitter}, low-resource intent identification and slot filling \cite{aepli-etal-2023-findings}. 
Moreover, machine translation techniques have been applied to re-write sentences from dialect to standard language \cite{kchaou-etal-2022-standardisation,pluss2020germeval,lambrecht-etal-2022-machine}. 
To overcome the limited availability of parallel training data, rule-based perturbations simulating dialectal morphosyntactic phenomena have been developed to generate synthetic parallel sentence pairs \cite{gerlach-etal-2022-producing}.

The emergence of pre-trained LMs has shifted the focus towards investigating disparities in representation and downstream performance between non-standard and standard languages.  
To this end, LM diagnostic tools encompass a wide range of techniques, including cloze tests \cite{zhang-etal-2021-sociolectal} and contrastive evaluation via minimal pairs \cite{demszky-etal-2021-learning}. 
\citet{ziems-etal-2022-value, ziems-etal-2023-multi} have created a rule-based translation system that converts English into various dialects. 
They use this system to conduct stress tests on multiple downstream models and reveal performance disparities between English dialects.

Methods to improve LMs' robustness towards dialects include integrating morphological information into LMs' tokenizers through inflection perturbations  \citep{tan-etal-2020-morphin, tan-etal-2020-mind},  manipulating the parse tree of the source sentence to align with the word order in the target dialect \cite{wang-eisner-2016-galactic,wu-etal-2023-oolong}, and character noise injection \cite{aepli-sennrich-2022-improving}. Using perturbed data during LM pre-training or adapter training has shown significant benefits for dialectal variants of the GLUE  tasks  \cite{wang-etal-2018-glue} specifically designed to dialects \cite{held-etal-2023-tada}.

A related line of research concentrates on the processing of spoken dialects, with a specific emphasis on dialectal speech to standard language recognition \cite{samardzic-etal-2016-archimob,pluss-etal-2022-sds} and spoken dialect identification \cite{zampieri-etal-2019-report}.

\section{Perturbations Based on Dialect Variations}
In this section, we introduce perturbations that are specifically motivated by dialectal variation. 
In English and German, these perturbations specifically focus on altering the morphosyntactic structure of the sentence to simulate dialects, while keeping the semantics unchanged.

\subsection{English Perturbations}
We re-use a set of perturbations obtained from the Multi-VALUE framework \cite{ziems-etal-2023-multi},\footnote{Usage terms at \url{https://value-nlp.org/}.} which translate text from Standard American English (SAE) to Urban African American Vernacular English (UAAVE). 
This set comprises a total of 118 perturbations, covering morphosyntactic phenomena present in UAAVE. 
The quality of the perturbation-based translation system is evaluated through prior human evaluation. 
These patterns are documented in and sourced from the Electronic World Atlas of Varieties of English (eWAVE, \citealp{ewave}), which lists 235 features from 75 English varieties, collected by 87 professional linguists in 175 peer-reviewed publications.

\subsection{German Perturbations} \label{subsec: german_perturb}
\begin{table*}[htp]
\begin{adjustbox}{max width=\textwidth, center}
\begin{tabular}{@{}lll@{}l@{}}
\toprule
\multicolumn{2}{l}{Category \qquad Perturbation} & Example: Before $\rightarrow$ After & Source  \\ \midrule
\multicolumn{2}{l}{\bf Noun Phrase}\\
\hspace{0.2em}
\perturbation
{possession\_von}
{\textit{von} construction instead of genitive}
{des Baums $\rightarrow$ von dem Baum}
{the\textsc{.gen} tree's $\rightarrow$ of the\textsc{.dat} tree}
{\citet{buelow2021structures};}
{\citet[map~77]{eichhoff4}}

\perturbation
{possession\_pron}
{Dative with poss.\ pron.\ instead of genitive}
{Kafkas Werke $\rightarrow$ Kafka seine Werke}
{Kafka's works $\rightarrow$ Kafka\textsc{.dat} his works}
{\citet{buelow2021structures};}
{\citet[map~77]{eichhoff4}}

\perturbation
{article\_name}
{Article before personal names}
{Franz Kafka $\rightarrow$ der Franz Kafka}
{Franz Kafka $\rightarrow$ the Franz Kafka}
{\citet{fleischer2019syntax};}
{\citet[map~76]{eichhoff4}}

\perturbation
{comparative}
{Comparitives with \textit{wie} or \textit{als wie}}
{größer als $\rightarrow$ größer wie}
{bigger than}
{\citet{jaeger2018vergleichskonstruktionen}}
{}

\perturbation
{double\_det}
{Emphatic double article}
{ein so großer Baum $\rightarrow$ ein so ein großer Baum}
{a such big tree $\rightarrow$ a such a big tree}
{\citet{auer2003nonstandard}}
{}

\midrule
\multicolumn{2}{l}{\bf Discourse \& Word Order} \\ 
\perturbation
{name\_order}
{Swapped family and given names}
{Franz Kafka $\rightarrow$ Kafka Franz}
{}
{\citet{auer2003nonstandard}}
{}

\perturbation
{denn}
{Obligatory particle \textit{denn} in questions}
{Wie ist das Wetter? $\rightarrow$ Wie ist denn das Wetter?}
{How is the weather? $\rightarrow$ How is \textsc{part} the weather?}
{\citet{fleischer2019syntax}}
{}

& \texttt{{verb\_clusters}} & {da sie das getan hat $\rightarrow$ da sie das hat getan} & {\citet{bader2009verb}}\\
& {Raised auxiliary/modal in 2-verb clusters} & \multicolumn{2}{l}{\gloss{because she it done had $\rightarrow$ because she it had done} \translation{because she had done it}} \\[2pt]%

\midrule
\multicolumn{2}{l}{\bf Tense \& Aspect} \\ 
\perturbation
{progressive}
{Progressive construction with \textit{am}}
{ich koche Suppe $\rightarrow$ ich bin Suppe am kochen}
{I cook soup $\rightarrow$ I am soup \textsc{prep} cooking}
{\citet{flick2013amprogressiv};}
{\citet{fleischer2019syntax}}

\midrule
\multicolumn{2}{l}{\bf Adverbs \& Prepositons} \\  
\perturbation
{pronominal\_adverbs}
{Splitting of pronominal adverbs with \textit{da-}}
{davon weiß ich nichts $\rightarrow$ da weiß ich nichts von}
{of.this know I nothing $\rightarrow$ there know I nothing of}
{\citet{fleischer2002spaltungskonstruktion}}

\perturbationoverlong
{direction}
{Directive preposition \textit{auf}}
{nach München $\rightarrow$ auf München}
{to Munich}
{\citet[p.~185]{merkle1993bairische};}
{\citeauthor{ada}}{(\citeyear{ada}, entry~\href{https://www.atlas-alltagssprache.de/r12-f4g/}{12/4g})\hspace{-5pt}}

\perturbation
{location}
{Locative preposition \textit{zu}}
{in München $\rightarrow$ zu München}
{in Munich}
{\citet[p.~186]{merkle1993bairische}}
{}

\midrule
\multicolumn{2}{l}{\bf Negation} \\ 
& \texttt{negative\_concord} & {ich sehe kein Haus $\rightarrow$ ich sehe kein Haus nicht} & {\citet{fleischer2019syntax, auer2003nonstandard}}\\
& {Negative concord} & \multicolumn{2}{l}{\gloss{I see no house $\rightarrow$ I see no house not} \translation{I don't see any house}}\\[3pt]

\midrule
\multicolumn{2}{l}{\bf Relativization} \\ 
\perturbation
{relative\_pron}
{Relative marker \textit{wo}}
{der Stern, der funkelt $\rightarrow$ der Stern, wo funkelt}
{the star \textsc{rel} sparkles}
{\citet{moser2023relative-particles}}
{}

\midrule
\multicolumn{2}{l}{\bf Complementation} \\ 
& \texttt{{es\_hat}} & {es gibt noch Brot $\rightarrow$ es hat noch Brot} & {\citet[p.~243]{koenig2015dtv}}\\
& {Existential clause \textit{es hat}} & \multicolumn{2}{l}{\gloss{it gives still bread $\rightarrow$ it has still bread} \translation{there is still bread left}} \\[2pt]%

\midrule
\multicolumn{2}{l}{\bf Verb Morphology} \\ 
\perturbation
{tun\_imperative}
{Periphrastic imperatives with \textit{tun} `do'}
{räum auf $\rightarrow$ tu aufräumen}
{tidy\textsc{.2sg.imp} up $\rightarrow$ do\textsc{.2sg.imp} tidy.up\textsc{.inf}}
{\citet[p.~66]{merkle1993bairische}}
{*}

\perturbation
{schwa\_elision}
{Schwa elision at the end of \textsc{1.sg.pres} verbs}
{ich habe $\rightarrow$ ich hab}
{I have}
{\citet{keel1980apocope}}
{}

\midrule
\multicolumn{2}{l}{\bf Pronouns} \\ 

\perturbation
{clitic\_es}
{Enclitic form of \textit{es} `it' after inflected verbs}
{ist es $\rightarrow$ ist's}
{is it}
{\citet{abraham1996personalpronomina}}
{}

\bottomrule
\end{tabular}%
\end{adjustbox}
\caption{%
Our collection of syntactic perturbations, sorted according to eWAVE's categories (in \textbf{bold face}). We give examples in German, with glosses in \gloss{gray italics}.
*This feature, \texttt{tun\_imperative}, is also inspired by systematic variation we could observe between the Standard and Swiss German versions of one of the datasets we use, xSID \citep{van-der-goot-etal-2021-masked, aepli-etal-2023-findings}.}
\label{tab:perturb-ex}
\end{table*}
\renewcommand{\arraystretch}{1}

Aligned with the Multi-VALUE framework, we implement a set of perturbations designed to translate text from Standard German into non-standard varieties.
Since there is \textit{no resource} detailing syntactic variations in German varieties similar to those for other languages such as English \citep{ewave}, North Germanic languages \cite{nordic-word-order-db}, creole and pidgin languages \cite{apics} or South American languages \cite{sails}, we review over thirty linguistic works published in the last decades.\footnote{While German dialectology has traditionally focused more on phonological/phonetic and lexical variation, we take advantage of the popularity that dialect syntax studies have gained in the past decades \citep[cf.][]{glaser1997dialektsyntax, scheutz2005perspektiven}.}
We select a set of morphosyntactic features that include different grammatical areas and features both regional and supraregional variation.
Similarly to the work by \citet{ziems-etal-2022-value}, our feature set is meant to showcase different types of variation rather than being exhaustive. 

\autoref{tab:perturb-ex} presents an overview of the perturbations, 
along with examples and pointers to relevant linguistic literature for further reference.\footnote{For a general introduction to syntactic variation in colloquial varieties of German, see \citet{fleischer2019syntax}.}
We group the perturbations according to eWAVE's category definitions and de-facto category assignments of similar English examples.\footnote{For instance, our \texttt{comparative} feature resembles eWAVE features \href{https://ewave-atlas.org/parameters/82}{82} and~\href{https://ewave-atlas.org/parameters/85}{85}.}
Several of our rules target grammatical areas that are not covered by eWAVE/\allowbreak{}Multi-VALUE, sometimes in ways relevant to the ToD context.
For instance, we also include changes to adpositions (relevant for labeling slots in queries relating to flight itineraries) and personal names (pertinent for queries like calling a contact or checking a birthday). 

We include features that are common and unmarked in colloquial German across all of the German-speaking area (such as eliding the word-final schwa in inflected verbs),
as well as some that are specific only to certain non-standard dialects (such as the choice of directive or locative preposition). 
Some of these features cannot be easily placed on this scale of regional specificity, as they might be licensed in more construction types in some areas than in others (like the progressive tense constructed with the preposition \textit{am}; \citealp{auer2003nonstandard}).
In total, we developed 18 perturbations that cover a wide range of phenomena.

\paragraph{Implementation.} Perturbation rules are implemented as rule-based functions that modify input sentences according to morphosyntax parses.
For part-of-speech (POS) tagging and dependency parsing, we employ German SoTA models in spaCy \cite{honnibal2020spacy} and Stanza \cite{qi-etal-2020-stanza}.
Noun inflection is handled using Derbi \cite{derbi}, verb conjugation is conducted with  Pattern-de\footnote{\href{https://digiasset.org/html/pattern-de.html}{digiasset.org/pattern-de}}  \cite{pattern}. 
We incorporate the list of first names from \citet{nett2019perceived}. Refer to \autoref{sec:examples} for examples of automatically perturbed sentences.

\begin{table*}[htp!]
    \centering
    \resizebox{0.99\textwidth}{!}{ %
    \begin{tabular}{ll@{\hspace{-18pt}}rl@{\hspace{-18pt}}rrccl}
    \toprule
    \textbf{Label}  & \textbf{Source} & \# \textbf{Langs.} &  \textbf{Domain}  & \# \textbf{Intents} & \# \textbf{Slots} & \textbf{Train / dev / test} & \textbf{DE tr?} & \textbf{License} \\ \midrule
    \xsid &  \makecell[tl]{\citet{van-der-goot-etal-2021-masked} \\ \citet{aepli-etal-2023-findings}} & 15 & General & 16 & 33 & 43k / 300 / 500 & & \href{https://bitbucket.org/robvanderg/xsid/src/master/LICENSE}{CC BY-SA 4.0}\\
    \atis & \citet{xu-etal-2020-end} & 9 & Aviasales & 18 & 84 & 3.7k / 1.2k / 893 & \cmark & \href{https://github.com/amazon-science/multiatis/blob/main/LICENSE}{Apache 2.0}\\
    \massive & \makecell[tl]{\citet{bastianelli-etal-2020-slurp} \\ \citet{fitzgerald-etal-2022-MASSIVE}} & 51 &  \makecell[tl]{Virtual assistant, \\ smart home} & 60 & 55 & 11k / 2k / 3k & \cmark & \href{https://github.com/alexa/massive/blob/main/LICENSE.txt}{Apache 2.0}\\
    \mtop & \citet{li-etal-2021-mtop} & 6 & Virtual assistant & 117 & 78 & 16k / 1.8k/ 3.5k & \cmark & \href{https://creativecommons.org/licenses/by-sa/4.0/}{CC BY-SA 4.0} \\
    \bottomrule 
    \end{tabular}%
    } 
    \caption{The datasets, used for experiments. Key: \textbf{\#~langs.} is the number of languages included in the dataset.  \textbf{\#~intents} and \textbf{\#~slots} stands for the the number of intents and slots in the dataset. \textbf{Train/dev/test} is the number of sentences in train, validation and test sets. \textbf{DE tr?} indicates whether training data in German is available. }
    \label{tab:datasets}
\end{table*}

\begin{table*}[htp!]
    \centering
    \resizebox{.95\textwidth}{!}{ %
    \begin{tabular}{lllcccl}
    \toprule
    \textbf{Label} & \textbf{HuggingFace ID} \cite{wolf-etal-2020-transformers} & \textbf{Source} & \# \textbf{Params.} & \textbf{Tr. data} & \textbf{Dialect?} & \textbf{License}\\ \midrule
    \mbert  & \href{https://huggingface.co/bert-base-multilingual-cased}{{ \texttt{bert-base-multilingual-cased}}} & \citet{devlin-etal-2019-bert} & 177M & Wiki &   \cmark  & \href{https://www.apache.org/licenses/LICENSE-2.0}{Apache 2.0}\\
    \xlmr   & \href{https://huggingface.co/xlm-roberta-base}{{  \texttt{xlm-roberta-base}}} & \citet{conneau-etal-2020-unsupervised} & 279M & CC &  & \href{https://opensource.org/license/mit/}{MIT}  \\
    \rembert & \href{https://huggingface.co/google/rembert}{{  \texttt{google/rembert}}} & \citet{chung2020rethinking} & 575M &  Wiki+CC  &  & \href{https://www.apache.org/licenses/LICENSE-2.0}{Apache 2.0}\\
    \mdeberta & \href{https://huggingface.co/microsoft/mdeberta-v3-base}{{ \texttt{microsoft/mdeberta-v3-base}}} & \citet{he2021debertav3,he2021deberta} & 276M  & CC &    & \href{https://opensource.org/license/mit/}{MIT} \\
    \cdashline{1-7}
    \mdistilbert & \href{https://huggingface.co/distilbert-base-multilingual-cased}{{  \texttt{distilbert-base-multilingual-cased}}} & \citet{sanh2019distilbert} & 134M  & Wiki  &  \cmark & \href{https://www.apache.org/licenses/LICENSE-2.0}{Apache 2.0}\\
    \mminilm & \href{https://huggingface.co/microsoft/Multilingual-MiniLM-L12-H384}{{  \texttt{microsoft/Multilingual-MiniLM-L12-H384}}} & \citet{wang2020minilm} & 117M  & CC & & \href{https://opensource.org/license/mit/}{MIT} \\
    \bottomrule
    \end{tabular}%
    } 
    \caption{The cross-lingual LMs used in the study.  Key: \textbf{Tr. data} denotes pre-training datasets, where Wiki stands for Wikipedia, CC stands for CommonCrawl \cite{wenzek-etal-2020-ccnet}.  \textbf{Dialect?} indicates whether German dialect data was explicitly included in the LM's pre-training data. The dashed line separates the base-size LMs from the distilled LMs. \mdistilbert is distilled from \mbert, \mminilm is distilled from \xlmr.}
    \label{tab:lms}
\end{table*}

\paragraph{Human evaluation.} 
We create a human evaluation dataset by manually labelling up to eight sentences per perturbation from each dataset. 
As certain rules can only be applied to fewer than eight sentences in some datasets, the human evaluation dataset comprises 200 sentences in total.

These sentences are assessed for fluency on a five-point Likert scale, where a score of 5 means that perturbed sentences are highly fluent and natural, while a score of 1 indicates the opposite. 
\autoref{sec:instruction} presents the annotation guidelines.
The annotations are carried out by two native German speakers with a background in computational linguistics and significant exposure to diverse dialects\footnote{One annotator is one of the authors. The second annotator was hired and received fair compensation according to the local employment regulations.}.

When evaluating the inter-annotator agreement based on raw scores, the percentage of cases where both annotators assign the same score is 53.51\% and the Pearson correlation coefficient is 0.51. 
Overall, the scores provided by both annotators average at 3.92 and 4.63.  
In 96 (48\%) and 3 (1.5\%) cases, both annotators give a score of 5 and 1 to the same sentence, respectively. 
Notably, the perturbations \texttt{verb\_clusters} shows significant disparity, with the mean score assigned by one annotator being~1, while the other annotator assigned a mean score of 5.\footnote{This feature is regionally very specific (\citealp{ada}, entry~\href{https://www.atlas-alltagssprache.de/runde-3/f13a-e/}{3/13abc}). The annotator providing high rankings is not from an area using this construction but was familiar with relevant literature and examples beforehand. The other annotator, unfamiliar until a pre-task explanation, gave lower rankings.
} 
Below is an example of a sentence pair that the annotators judged with opposite scores (1 vs.\ 5). \textit{A} is for German, \textit{B} is for the dialect re-write. 
The fragment of the sentence affected with the \texttt{verb\_clusters} perturbation is underlined.

\noindent
\begin{tabular}{@{}l@{\quad}l@{ }l@{ }l@{ }l@{ }l@{ }l@{ }l@{}}
\textit{A} & Frag & ob & Pauline & zu & meinem & Thanksgiving & - \\
& \textit{Ask} & \textit{if} & \textit{Pauline} & \textit{to} & \textit{my} & \textit{Thanksgiving} & \textit{-}
\end{tabular}
\noindent
\begin{tabular}{@{}l@{\quad}l@{ }l@{ }l@{ }l@{ }l@{ }l@{ }l@{ }l@{ }l@{ }l@{}}
\phantom{A}&Treffen & \textcolor{purple}{\underline{kommen}} & \textcolor{green}{\underline{will}} &  . \\
&\textit{gathering} & \textcolor{purple}{\textit{come}.\textsc{inf}} & \textcolor{green}{\textit{wants}} & .
\end{tabular}

\noindent
\begin{tabular}{@{}l@{\quad}l@{ }l@{ }l@{ }l@{ }l@{ }l@{ }l@{}}
\textit{B} & Frag & ob & Pauline & zu & meinem & Thanksgiving & - \\
& \textit{Ask} & \textit{if} & \textit{Pauline} & \textit{to} & \textit{my} & \textit{Thanksgiving} & \textit{-}
\end{tabular}
\noindent
\begin{tabular}{@{}l@{\quad}l@{ }l@{ }l@{ }l@{ }l@{ }l@{ }l@{ }l@{ }l@{ }l@{}}
\phantom{A}&Treffen & \textcolor{green}{\underline{will}} &  \textcolor{purple}{\underline{kommen}}  & . \\
&\textit{gathering} & \textcolor{green}{\textit{wants}} &  \textcolor{purple}{\textit{come}.\textsc{inf}}  & . 
\end{tabular}

Similar discrepancies are observed in other perturbations such as \texttt{pronominal\_adverbs}, \texttt{relative\_pron}, and \texttt{name\_order}.

Additionally, we map the score to a binary scale (where scores 1 and~2 were grouped as~0, and scores 3, 4, and~5 were grouped as~1). 
The exact match agreement  becomes 91.89\%. Cohen's kappa \cite{mchugh2012interrater} reaches a 0.61. 
Areas of disagreement include \texttt{verb\_clusters}  and \texttt{progressive}. 
These perturbations account for the majority of the discrepancies, with 7 items and 4 items respectively.
The results indicate moderate to substantial levels of agreement between annotators and shed light on which perturbations tend to cause the most disagreement. 
Since linguistic acceptability in the context of language variation can be subjective, we chose to keep all perturbations, even if there were disagreements among annotators.

\section{Methodology} \label{sec:method}

We choose task-oriented dialogue systems as a task where we expect a high level of linguistic variation in real-life application settings. 
There is limited research on whether these systems commonly encounter inputs from dialect speakers in real-world applications \cite{bird-2020-decolonising,nekoto-etal-2020-participatory}. 
Nevertheless, several works encourage the localization of dialogue systems to dialect varieties. 
One common motivational aspect shared by these works is the aim to encourage the use of dialects, with the expectation of positively impacting the prestige of the language \cite{trong2019enabling,aepli-etal-2023-findings}.

\paragraph{Datasets.} \autoref{tab:datasets} provides a brief description of the ToD datasets for intent recognition and slot filling. All of the datasets considered support zero-shot cross-lingual setups by including English training and German development and test data. 
Except for \xsid, all datasets are further equipped with German training data.
In this study, we concentrate on German and English, leaving other languages for future work.

\paragraph{Method.} We adopt a joint approach for intent detection and slot filling, leveraging the implementation of MaChAmp  \cite{van-der-goot-etal-2021-massive}. 
It uses an encoder and a separate decoder head for each task, one for intent classification and one for slot detection with a CRF layer on top.
We use the default settings, which include a learning rate of 0.0001.
We experiment with six encoder-based multilingual LMs (\autoref{tab:lms}).
Each LM undergoes training with five random seeds, and results are averaged across all runs. 
LMs are trained on a single NVIDIA A100 device.

\begin{table*}[h]
\begin{center}
    
\resizebox{0.95\textwidth}{!}{ %
\begin{tabular}{llcccccccc}
\toprule
& &  \multicolumn{2}{c}{Intact} &  \multicolumn{2}{c}{Individual Perturbations} &  \multicolumn{2}{c}{All Perturbations} \\  
&   &   Intent Acc  &  Slot F$_{1}$ &  $\Delta$ Intent Acc  & $\Delta$ Slot F$_{1}$ & $\Delta$  Intent Acc & $\Delta$   Slot F$_{1}$  \\  
\cmidrule(lr){3-4}  \cmidrule(lr){5-6}  \cmidrule(lr){7-8}
\xsid & \mbert & 76.36 & 70.57 & 0.40 & 2.32 & 5.60 & 20.70 \\
 & \xlmr & 90.20 & 76.23 & 0.31 & 2.70 & 4.08 & 22.95 \\
 & \rembert & 91.08 & 79.44 & 0.34 & 2.78 & 4.16 & 23.59 \\
 & \mdeberta & 94.88 & 82.62 & 0.24 & 2.69 & 3.12 & 23.03 \\ \cdashline{2-8}

 & \mdistilbert & 71.04 & 66.62 & 0.43 & 2.17 & 4.88 & 19.94 \\
 & \mminilm & 72.16 & 69.29 & 0.34 & 2.25 & 3.56 & 22.36 \\ \midrule
\atis & \mbert  & 76.91 & 62.22 & 0.07 & 2.50 & 0.81 & 9.57 \\
 & \xlmr & 78.75 & 76.18 & 0.02 & 3.72 & 0.18 & 11.13 \\
 & \rembert & 79.28 & 83.32 & 0.01 & 4.05 & 0.27 & 15.95 \\
 & \mdeberta & 79.17 & 80.10 & 0.01 & 3.89 & 0.27 & 10.93 \\ \cdashline{2-8}
 & \mdistilbert  & 74.67 & 56.72 & 0.05 & 2.38 & 0.43 & 8.74 \\
 & \mminilm & 74.65 & 68.49 & 0.00 & 3.12 & 0.25 & 9.21 \\ \midrule

\massive & \mbert  & 54.63 & 49.25 & 0.43 & 2.38 & 5.74 & 21.56 \\
 & \xlmr & 74.86 & 65.75 & 0.42 & 2.80 & 6.70 & 26.47 \\
 & \rembert & 83.86 & 73.33 & 0.41 & 3.02 & 6.29 & 27.64 \\
 & \mdeberta & 83.91 & 73.86 & 0.39 & 3.02 & 6.29 & 28.08 \\ \cdashline{2-8}
 & \mdistilbert & 45.53 & 42.74 & 0.38 & 1.99 & 4.42 & 19.30 \\
 & \mminilm & 58.14 & 54.57 & 0.30 & 2.44 & 5.34 & 23.02 \\ \midrule

\mtop & \mbert  & 67.34 & 66.99 & 0.51 & 2.58 & 8.20 & 26.96 \\
 & \xlmr & 88.76 & 77.53 & 0.60 & 2.96 & 8.88 & 30.44 \\
 & \rembert  & 91.35 & 79.33 & 0.58 & 3.05 & 8.79 & 31.41 \\
 & \mdeberta & 90.66 & 79.26 & 0.60 & 2.95 & 8.24 & 30.50 \\ \cdashline{2-8}
 & \mdistilbert & 58.72 & 59.71 & 0.46 & 2.50 & 7.32 & 25.79 \\
 & \mminilm & 75.89 & 70.53 & 0.52 & 2.79 & 7.17 & 29.13 \\

\midrule
Mean &  & 76.37 & 69.36 & 0.33 & 2.79 & 4.62 & 21.60 \\

\bottomrule
\end{tabular}%
 } 
\end{center}

\caption{The overall results for intent recognition and slot filling on test sets in German in zero-shot setup \textit{(i)} and the gap in performance before and after dialect perturbations are applied (in percentage points). Intact (left): performance on intact test sets. Individual perturbations (middle):  18 individual perturbations are applied and average performance gap is computed across them. All perturbations (right): all perturbations applied simultaneously. $\Delta$ denotes the difference between performance on intact and perturbed data. Performance on intact data consistently surpasses that on perturbed data, leading to positive $\Delta$ values. The results are averaged across five runs with varying random initialization. }
\label{tab:dialect_results}
\end{table*}

\paragraph{Experimental setup.} Evaluation metrics are accuracy for intent recognition and the span F$_1$ score for slot filling, where both span and label must match exactly.
We explore three experimental setups: 
\begin{inparaenum}[\itshape (i)]
  \item zero-shot setup: models are trained on English training data;
  \item zero-shot setup with German development data;
  \item fully supervised setup (where available): models trained on German training data. 
\end{inparaenum}

Model selection over epochs is based on its performance on development data in English~\textit{(i)} and German \textit{(ii, iii)}, without any access to labeled UAAVE or German data during the training phase.

To assess the robustness of the ToD model, we apply perturbations to generate synthetic UAAVE and German dialect test data. 
We then use fine-tuned models to make predictions on this perturbed data. 
We evaluate the impact of these perturbations by measuring \textit{the difference in performance before and after} the perturbation is applied.
In addition, following the research on adversarial attacks \cite{tsai-etal-2019-adversarial-attack} we define the \textit{success rate} of a perturbation as the number of instances that become misclassified after the perturbation was applied.

\section{Results} \label{sec:results}

\paragraph{RQ~1: What is the impact of perturbed data on performance?} \autoref{tab:dialect_results} and \autoref{tab:UAAVE_results} (\autoref{sec:tables})  present the intent recognition and slot filling test results for zero-shot \textit{(i)} German and English, respectively, with and without perturbations. 
Additionally, in \autoref{sec:tables}, \autoref{tab:ende_results} displays the results for setup \textit{(ii)}, while \autoref{tab:dede_results} presents the fully-supervised German setup \textit{(iii)}.
The performance scores align with earlier results reported in the dataset papers and recent research \cite{aepli-etal-2023-findings}. 
The perturbations are used in two scenarios:
\begin{inparaenum}[\itshape (a)] 
\item with 18 German and 118 English perturbations applied individually and average performance computed across them,\footnote{While some of the syntactic features tend to co-occur, e.g., the \texttt{name\_order} swap is most commonly found in varieties that also exhibit the \texttt{article\_name} feature (\citealp{ada}, entry~\href{https://www.atlas-alltagssprache.de/r10-f16ab/}{10/16ab}). We nevertheless apply rules individually in scenario~\textit{(a)}, as the borders between feature areas do not form perfect isoglosses.
In the given example, name swapping without any added article is attested in some locations near the Belgian and Dutch borders (ibid.).} 
\item with all perturbations applied simultaneously.
\end{inparaenum}

\autoref{tab:dialect_results} shows the performance gap\footnote{All of the performance changes detailed in the following are in percentage points.} in zero-shot evaluation on test sets before and after German perturbations are applied concerning the dataset and the LM. 
The decrease in performance is minimal for intent recognition accuracy, averaging at 0.33, when individual perturbations are applied. 
However, it drops further by an average of 4.62 when all perturbations are applied simultaneously. 
The drop is more pronounced for slot filling, where performance  decreases by 2.79 Slot $F_1$ after individual perturbations and by 21.60 Slot $F_1$ after the simultaneous application of all perturbations. 

In the evaluation for English (\autoref{tab:UAAVE_results}, \autoref{sec:tables}), we observe similar trends. 
The decline in intent recognition is minimal, with average drops of merely 0.10 up to 2.48 accuracy in the two considered scenarios. 
Conversely, the decline in slot filling is more pronounced, with 9.87 and 49.37 F$_1$ score on average for individual and combined perturbations, respectively. 
The simultaneous application of all perturbations affects the performance more than applying individual perturbations.

Further experiments with setup~\textit{(ii)} show that the choice between English or German development data has no significant impact on the performance on perturbed data (compare \autoref{tab:dialect_results} with \autoref{tab:ende_results}, \autoref{sec:tables}). In particular, while zero-shot downstream performance improves for all LMs (e.g.\ \mdeberta and \rembert, show gains of 0.61 accuracy  and 0.18 F$_1$ score  and 0.77 accuracy  and 2.36 F$_1$ score, respectively),  the impact of the perturbations remains similar with comparable results to the results discussed earlier        in the setup~\textit{(ii)} (higher impact on slots than intents).

In the fully-supervised setup~\textit{(iii)} with fine-tuning on German data (\autoref{tab:dede_results}, \autoref{sec:tables}), we observe an expected significant improvement in performance across all three datasets, due to the in-language training data.
While the performance drop is almost identical to the zero-shot set-up for intent accuracy, the slot filling performance is considerably more robust. Here, the average drop is only 6.27~F$_1$ when all perturbations are applied (compared to 21.60 in the zero-shot set-up).
This suggests that fine-tuning with in-language data improves performance on both intact and perturbed test sets.

To sum up, while LMs can still produce accurate predictions on the sentence level after the sentence is perturbed with dialectal variations (i.e., intent recognition), their performance suffers particularly on the word level (i.e., slot filling), and this becomes more pronounced as the sentence's perturbation increases. 
Fine-tuning with in-language data improves overall performance and enhances significantly the treatment of perturbed data.
These findings remain consistent across all four datasets and the various LMs considered.

\begin{figure}
    \centering
    \includegraphics[width=\columnwidth]{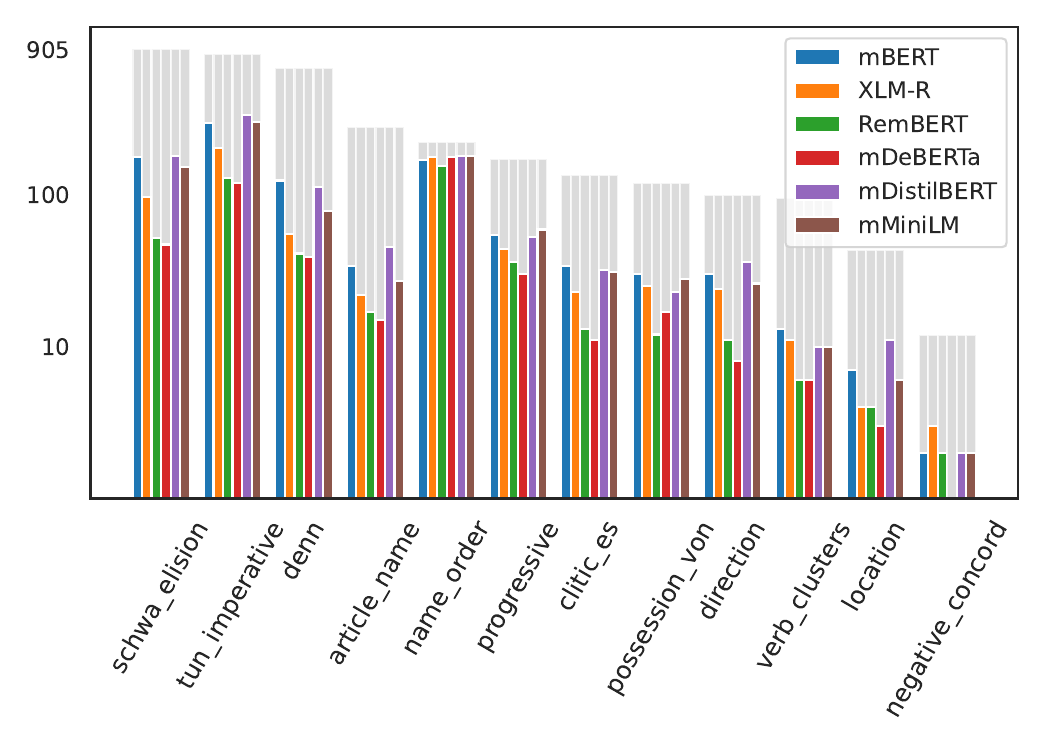}
    \caption{Intent prediction success rates on the perturbed German test set on \massive with respect to most impactful individual perturbations. The grey bars denote the count of perturbed sentences,  the colored bars show the success rate. 
    A logarithmic scale is used.}
    \label{fig:MASSIVE_success_rate}
\end{figure}
 
\paragraph{RQ~2: Which perturbations affect performance the most?}   This part focuses on the zero-shot scenario~\textit{(i)}.
First, we examine perturbations that result in a non-zero perturbation success rate, indicating their ability to change the predicted intent.
\autoref{fig:MASSIVE_success_rate} illustrates the success rate of 12 individual perturbations on the German test set of \massive, compared with the count of perturbed sentences. 
The six remaining perturbations do not affect the performance and have zero success rate.
While all perturbations preserve semantics, those with higher success rates induce a more substantial shift in the representation space and effectively fool LMs. 
The perturbations \texttt{schwa\_elision} and \texttt{tun\_imperative} impact a similar number of sentences, yet their success rates differ, with the latter exhibiting a higher success rate. 
This could be attributed to the alteration in the number of words in \texttt{tun\_imperative} and the change in the position of the main content word, shifting from the first to the final position in the sentence (see the example in \autoref{fig:example}).
The \texttt{name\_order} perturbation exhibits the highest success rate, while the \texttt{negative\_concord} perturbation demonstrates the lowest non-zero success rate.
The analysis of success rates in German and English across various datasets (Figures \ref{fig:en_success_rate} and \ref{fig:de_success_rate}, \autoref{sec:success_rate}) confirms that the frequency of perturbations differs across datasets due to their design.
However, the success rates remain consistent.
There are frequent perturbations that have little impact, such as \texttt{location} and \texttt{direction} in German (except for \atis{}, see below), and \texttt{zero\_plural} in English.
Some perturbations demonstrate consistently stable success rates in all four datasets, as observed in the case of \texttt{progressive} in English and \texttt{word\_order} in German.
This could be linked to the frequency of respective dialect phenomena in the LM's pre-training data, where rarely seen dialect phenomena deceive it more effectively.

\begin{figure}
    \centering
    \includegraphics[width=\columnwidth]{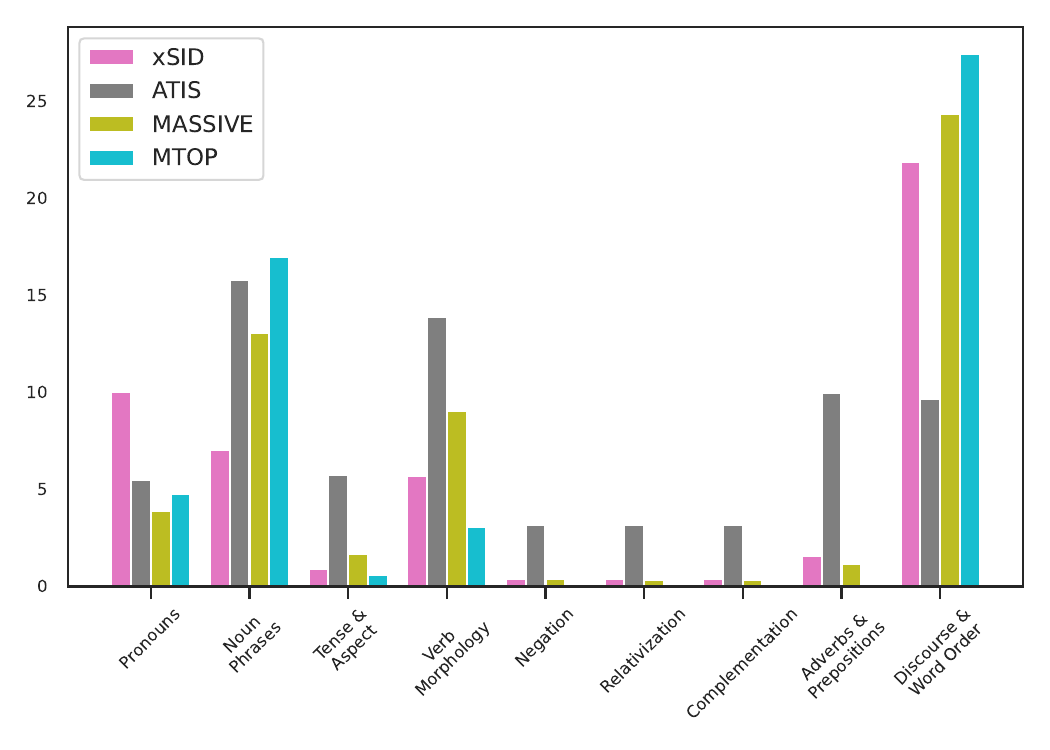}
    \caption{The $\Delta$ slot F$_1$ score of the best performing \mdeberta with respect to perturbation category in perturbed German test set in four datasets. $\Delta$ denotes the difference in F$_1$ score between performance on intact and perturbed data. }
    \label{fig:f1_drop_cat}
\end{figure}
\autoref{fig:f1_drop_cat} examines how the F$_1$ score declines after individual perturbations are applied. 
Here, the perturbations are grouped according to eWAVE categories, and \mdeberta serves as the backbone LM.
Across datasets, the F$_1$ score is mostly affected by the three perturbations falling under the \textbf{Discourse \& Word Order} category, followed by perturbations affecting \textbf{Noun Phrases} and \textbf{Verb Morphology}. %
In turn, in English the \textbf{Tense \& Aspect} category stands out, followed by \textbf{Pronouns} and \textbf{Noun Phrases} (\autoref{sec:f1_drop_cat}).

There are  structural and domain-specific variations in performance across datasets.
In \xsid, the \textbf{Pronouns} category experiences a significant impact, indicating a higher frequency of the usage of \textit{es} `it' (shortened to \textit{'s} by our perturbations) compared to other datasets. 
In \atis, the \textbf{Adverbs \& Prepositions} category is notably affected. 
This category includes perturbations that modify directive and locative prepositions, which are commonly employed in \atis due to its specific domain (with queries like ``What are flights \textit{to} X that also stop \textit{in} Y?'').

\paragraph{RQ~3: How does the performance of LMs vary?} 
\autoref{tab:dialect_results} shows that  in the zero-shot setup \mdeberta consistently outperforms other LMs, followed closely by \rembert.
\xlmr and \mbert also exhibit competitive performance, while \mdistilbert and \mminilm tend to have lower scores.
There is a consistent drop in performance  when dialect perturbations are applied, indicating that all LMs are sensitive to dialectal variations.
\autoref{fig:MASSIVE_success_rate} exhibits similar trends across all LMs, with \mdeberta and \rembert displaying comparatively lower success rates for individual perturbations. 
Conversely, distilled models, \mdistilbert and \mminilm, show higher success rates.

Our results  suggest that \mdeberta and \rembert are more robust to dialectal variations,  outperforming other LMs in both tasks across four datasets. 
This aligns with previous cross-lingual studies \cite{adelani-etal-2022-masakhaner,malmasi-etal-2022-semeval}, where they outperformed other LMs and demonstrated superior results in lower-resource settings.

\paragraph{Error analysis.} Next, we focus on German for error analysis. In intent recognition, LMs often confuse semantically similar intents (\textsc{PlayMusic} and \textsc{SearchCreativeWork}), or intents associated with the same service, (\textsc{alarm/\allowbreak{}cancel\_alarm} and \textsc{alarm/\allowbreak{}set\_alarm}, \xsid). These errors become apparent when the LMs are tested on intact data and become even more pronounced when dialect perturbations are applied. Lastly, LMs tested on perturbed data tend to misinterpret intents that commonly share homonymous words (\textsc{BookRestaurant} and \textsc{RateBook}, \xsid).

There are three primary errors in slot filling. Firstly, the LMs incorrectly identify slot boundaries when perturbations impact word order. In such cases, the LM tends to make errors in predicting slot boundaries, as observed in instances like ``Merkel Angela'' (\textsc{B-Person I-Person}) transformed from ``Angela Merkel'', where the LMs often predict \textsc{B-Person B-Person}, splitting the span inaccurately. Secondly, when the word order is maintained, the LMs exhibit more mistakes in predicting slot types. For instance, when the direction perturbation is applied, the LMs frequently assign incorrect slot types. Finally, when an extra auxiliary verb is introduced, as in the case of the \texttt{progressive} perturbation, LMs frequently assign it a slot label.

\section{Conclusion and Future Work}

This project tests the robustness of task-oriented dialogue systems (ToD) towards English and German dialects.  
Our methodology involves applying rule-based perturbations to translate ToD datasets from Standard American English to Urban African American Vernacular English, and from German to its non-standard variety. 
To the best of our knowledge, we are the first to design such perturbations for German.
Subsequently, we train multiple joined ToD models, equipped with various Transformer-based backbones, assessing their performance on intact and perturbed data.

We conclude, that \begin{inparaenum}[\bf Re~RQ~1:]
\item The impact of perturbed data on LM performance varies depending on the type of perturbation and the task. In general, we note a minor decrease in intent recognition  but a notable drop in slot filling. Issues in slot filling involve inaccuracies in boundary identification, mistakes in predicting slot types with altered word order, and frequent misalignments of slot labels with an extra verb.
\item Across languages, the performance drop varies by dataset and LM, indicating domain and language-specific patterns in response to phenomena-based perturbations.
\item There is no clear winner, but \mdeberta and \rembert outperform other LMs by gaining higher performance scores and being more robust to dialectal variations.
\end{inparaenum}

Future work includes \begin{inparaenum}[(i)] 
\item extension to other languages with distinct dialectal variation; 
\item development of fair evaluation approaches, that do not favor standard languages but account for dialects;
\item incorporating phonological phenomena for a deeper understanding of dialectal variations in  written and spoken forms.
\end{inparaenum}

Conducting similar experiments with other languages and dialects can help in understanding how these models generalize across diverse linguistic landscapes.

\section*{Limitations}

\noindent \textbf{Focus on written text.}  Our study  predominantly focuses on written text, and we do not account for phonological or lexical differences between the standard language and non-standard varieties. Our emphasis is primarily on syntactic differences, and as such, we acknowledge that our analysis may not fully capture the complete spectrum of linguistic nuances present in spoken language variation. 

\noindent \textbf{Choice of LMs.} Our choice of LMs is inherently limited; we do not use auto-regressive or sequence-to-sequence language models for the sake of compute time.

\noindent \textbf{German perturbations.} The selected German perturbations do not perfectly capture any particular German dialect, but they are based on prevalent patterns found in a selection of dialects and colloquial varieties.

\noindent \textbf{Design of perturbations.} The perturbation rules, borrowed from \citet{ziems-etal-2023-multi} for English and developed by us for German, specifically target syntactic phenomena, excluding orthographic and lexical variations.

\noindent \textbf{Focus on zero-shot settings.} In our approach, the primary focus is on zero-shot settings, where dialect data is intentionally excluded from the training process to prevent any potential leakage. This choice allows us to follow a practical scenario where the model can handle diverse dialects without the need for collecting specific dialect data during training. However, deviating from the zero-shot setting could potentially yield models that are more robust to direct perturbation. In such cases, the upper bound for evaluating robustness would involve incorporating dialect training data, providing an alternative perspective to the zero-shot approach.

\section*{Ethical considerations}

\noindent \textbf{Human assessment.} 
This work involves human assessment of synthetically generated data. Two annotators were involved. One annotator is one of the authors. The second annotator
was hired and received fair compensation according to the local employment regulations.

\noindent \textbf{Perturbation rules.}
Our software allows automatically applying changes to German sentences that simulate dialectal and colloquial variation.
Our selection of perturbation rules is not exhaustive enough to simulate any one dialect and is taken to be representative of the breadth of variation in the German dialect landscape.
Because of these restrictions, we find it unlikely that our system could be used for the mockery and parody of any dialects or registers.
We release the code for perturbations for research purposes only and expressly forbid usage for mockery or parody of any dialects or registers.

\section*{Acknowledgements}
We thank our colleagues from the MaiNLP research group and the anonymous reviewers for their feedback.
This research is supported by the ERC Consolidator Grant DIALECT 101043235.

\bibliography{ling,anthology,mypaper}

\newpage
\clearpage

\appendix
\onecolumn 

\section{Examples of perturbed sentences} \label{sec:examples}

\begin{table*}[htp!]
\centering
\begin{adjustbox}{max width=\textwidth, center}
\begin{tabular}{lp{7.1cm}p{7cm}}
\toprule
 \textbf{Perturbation} & \textbf{Sentence} \hspace{4cm} $\rightarrow$ & \textbf{Perturbed sentence} \\
\midrule

\multicolumn{3}{l}{\bf Noun Phrase}\\

\texttt{possession\_von} & Welcher Ort steht in der Erinnerung für das Abendessen des Schachclubs ?	& Welcher Ort steht in der Erinnerung für das Abendessen vom Schachclub ?  \\
& \translation{What's the location of the Chess Club dinner reminder?}\\
 
\texttt{possesion\_pron} & Wann ist Robin Williams Geburtstag ? & Wann ist Robin William sein Geburtstag ? \\
& \translation{What is Robin William's birthday?} \\

\texttt{article\_name} & Email an Natalie zu ihrem Geburtstag .	& Email an die Natalie zu ihrem Geburtstag . \\
& \translation{Email Natalie for her birthday.}\\

\texttt{comparative} & Wird es morgen heißer als 25 Grad Celsius ? & 	Wird es morgen heißer wie 25 Grad Celsius ? \\
& \translation{Will it be hotter than 25°C?} \\

\texttt{double\_det} & Ich möchte noch ein so lustiges Lied hören & Ich möchte noch ein so ein lustiges Lied hören \\
& \translation{I want to hear another song this funny} \\

\multicolumn{2}{l}{\bf Discourse \& Word Order}\\

\texttt{name\_order} & Ruf stattdessen Gloria Burgess an &	Ruf stattdessen Burgess Gloria an \\
& \translation{Call Gloria Burgess instead} \\

\texttt{denn} &  Wie lange geht meine \textit{[sic]} Timer noch ?	& Wie lange geht denn meine Timer noch ? \\
& \translation{How much time is left on my timer?}\\

\texttt{verb\_clusters} & Zeige alle Erinnerungen an , die mit Familie zu tun haben . &  Zeige alle Erinnerungen an , die mit Familie haben zu tun  . \\
& \translation{Show all family reminders}\\

\multicolumn{2}{l}{\bf Tense \& Aspect}\\

\texttt{progressive} & Ich höre Jazz  . & Ich bin Jazz am hören .\\
& \translation{I listen to jazz.}\\

\multicolumn{2}{l}{\bf Adverbs \& Prepositions}\\

\texttt{pronominal\_adverbs} &  Stelle dafür einen Timer . & Stelle da einen Timer für .\\
& \translation{Set a timer for this.}\\

\texttt{direction} &  Berechne eine Route nach Hamburg . &	Berechne eine Route auf Hamburg . \\
& \translation{Calculate the route to Hamburg.}\\

\texttt{location} & Was kostet der Bodentransport in Denver ? & Was kostet der Bodentransport zu Denver ? \\
& \translation{How much is ground transportation in Denver?}\\

\multicolumn{2}{l}{\bf Negation}\\

 \texttt{negative\_concord} &  Nimm heute keine Anrufe an . & 	Nimm heute keine Anrufe nicht an . \\
& \translation{Don't take any calls today.}\\

\multicolumn{2}{l}{\bf Relativization}\\

\texttt{relative\_pron} &  Freunde , die jetzt online sind	& Freunde , wo jetzt online sind \\ 
& \translation{Friends who are online right now}\\

\multicolumn{2}{l}{\bf Complementation}\\ 

 \texttt{es\_hat} &  Sende Andre die neuesten IT Themen die es gibt &	Sende Andre die neuesten IT Themen die es hat . \\

\multicolumn{2}{l}{\bf Verb Morphology}\\ 

\texttt{tun\_imperativ} & Erinnere mich an notwendige veranstaltungen . & 	Tu mich an notwendige veranstaltungen erinnern .\\
& \translation{Remind me of necessary events.}\\

\texttt{schwa\_elision} & Welche Erinnerungen habe ich für meinen Chef ? & 	Welche Erinnerungen hab' ich für meinen Chef ?\\ 
& \translation{What reminders do I have for my boss?}\\

\multicolumn{2}{l}{\bf Pronouns}\\ 

\texttt{clitic\_es} &  Wird es für die Party am Samstag sonnig ? &	Wird's für die Party am Samstag sonnig ? \\
& \translation{Will it be sunny for the party on Saturday?}\\

\bottomrule

\end{tabular}
\end{adjustbox}
\caption{Examples of automatically perturbed sentences from the task-orieneted datasets used in this study.}
\label{tab:perturbed_sentences}
\end{table*}

\newpage
\clearpage

\section{Performance in joint intent recognition and slot filling} \label{sec:tables}

\subsection{Performance on perturbed English test sets}

\begin{table*}[h]
\resizebox{0.95\textwidth}{!}{ %
\begin{tabular}{llcccccc}
\toprule
& &  \multicolumn{2}{c}{Intact} &  \multicolumn{2}{c}{Individual Perturbations} &  \multicolumn{2}{c}{All Perturbations} \\  
&   &   Intent Acc  &  Slot F$_{1}$ &  $\Delta$ Intent Acc  & $\Delta$ Slot F$_{1}$ & $\Delta$  Intent Acc & $\Delta$   Slot F$_{1}$  \\  
\cmidrule(lr){3-4}  \cmidrule(lr){5-6}  \cmidrule(lr){7-8}

\xsid & \mbert & 99.04 & 95.28 & 0.09 & 10.34 & 2.32 & 57.14 \\
 & \xlmr & 99.20 & 95.93 & 0.11 & 9.94 & 1.96 & 57.66 \\

 & \rembert & 99.12 & 96.11 & 0.04 & 9.90 & 1.32 & 57.68 \\
 & \mdeberta & 99.04 & 96.00 & 0.07 & 9.96 & 1.68 & 58.01 \\ \cdashline{2-8}
 & \mdistilbert & 99.00 & 94.52 & 0.09 & 10.19 & 2.36 & 56.61 \\
 & \mminilm & 99.24 & 95.20 & 0.06 & 9.74 & 1.84 & 58.11 \\ \midrule

\atis & \mbert & 79.69 & 93.00 & 0.00 & 10.98 & 0.04 & 45.46 \\
 & \xlmr & 79.75 & 92.99 & 0.00 & 10.82 & 0.18 & 45.30 \\
 & \rembert & 79.73 & 93.31 & 0.00 & 11.04 & 0.18 & 45.59 \\
 & \mdeberta & 79.84 & 92.93 & 0.00 & 10.76 & -0.07 & 46.31 \\ \cdashline{2-8}
 & \mdistilbert & 79.78 & 92.98 & 0.00 & 10.99 & 0.07 & 45.43 \\
 & \mminilm & 75.39 & 90.76 & 0.00 & 10.79 & -0.04 & 44.24 \\ \midrule

\massive & \mbert & 87.95 & 81.92 & 0.14 & 14.08 & 4.93 & 44.50 \\
 & \xlmr & 89.11 & 82.79 & 0.17 & 13.78 & 3.88 & 44.42 \\
 & \rembert & 89.25 & 83.10 & 0.08 & 14.21 & 3.57 & 44.82 \\
 & \mdeberta & 89.59 & 82.99 & 0.24 & 14.32 & 2.68 & 44.64 \\ \cdashline{2-8}
 & \mdistilbert & 87.11 & 80.65 & 0.27 & 14.17 & 5.20 & 43.61 \\
 & \mminilm & 84.77 & 79.91 & 0.15 & 13.65 & 4.21 & 43.14 \\ \midrule

\mtop & \mbert & 96.40 & 89.14 & 0.18 & 4.54 & 4.83 & 50.58 \\
 & \xlmr & 96.65 & 89.78 & 0.10 & 4.58 & 3.16 & 50.46 \\
 & \rembert & 97.15 & 89.83 & 0.13 & 4.52 & 4.17 & 50.44 \\
 & \mdeberta & 96.71 & 89.24 & 0.11 & 4.43 & 3.07 & 49.98 \\ \cdashline{2-8}
 & \mdistilbert & 96.01 & 88.53 & 0.19 & 4.56 & 4.52 & 50.19 \\
 & \mminilm & 93.17 & 88.84 & 0.18 & 4.53 & 3.45 & 50.47 \\ 

\midrule
Mean &  & 90.53 & 89.82 & 0.10 & 9.87 & 2.48 & 49.37 \\
\bottomrule
\end{tabular}%
 } 
\caption{The overall results for intent recognition and slot filling on test sets in  English and the gap in performance before and after UAAVE perturbations are applied. Intact (left): performance on intact test sets. Individual perturbations (middle):  118 individual perturbations are applied and average performance gap is computed across them. All perturbations (right): all perturbations applied simultaneously. $\Delta$ denotes the difference between performance on intact and perturbed data. Performance on intact data consistently surpasses that on perturbed data, leading to positive $\Delta$ values. The results are averaged across five runs with varying random initialization. }
\label{tab:UAAVE_results}
\end{table*}

\newpage
\clearpage

\subsection{Performance with German development set on perturbed German test sets}

\begin{table*}[h]
\resizebox{0.95\textwidth}{!}{ %
\begin{tabular}{llcccccc}
\toprule
&    &  \multicolumn{2}{c}{Intact}    &  \multicolumn{2}{c}{Individual Perturbations} &  \multicolumn{2}{c}{All Perturbations} \\  
&   &   Intent Acc  &  Slot F$_{1}$ &  $\Delta$ Intent Acc  & $\Delta$ Slot F$_{1}$ & $\Delta$  Intent Acc & $\Delta$   Slot F$_{1}$  \\  
\cmidrule(lr){3-4}  \cmidrule(lr){5-6}  \cmidrule(lr){7-8}

\xsid & \mbert & 78.72 & 71.81 & 0.43 & 2.47 & 6.12 & 21.28 \\
  & \xlmr & 91.08 & 78.19 & 0.26 & 2.76 & 3.44 & 23.46 \\
  & \rembert & 94.88 & 83.12 & 0.39 & 2.78 & 5.20 & 23.76 \\
  & \mdeberta & 96.88 & 83.08 & 0.27 & 2.73 & 3.28 & 23.15 \\ \cdashline{2-8}
  & \mdistilbert & 75.88 & 66.33 & 0.58 & 2.23 & 6.24 & 19.78 \\
  & \mminilm & 72.32 & 70.51 & 0.31 & 2.39 & 2.84 & 22.64 \\ \midrule
\atis & \mbert & 76.89 & 62.73 & 0.05 & 2.57 & 0.56 & 9.46 \\
 & \xlmr & 79.08 & 78.49 & 0.02 & 3.88 & 0.27 & 11.52 \\
 & \rembert & 79.24 & 83.82 & 0.02 & 4.11 & 0.36 & 11.03 \\
 & \mdeberta & 78.84 & 80.12 & -0.01 & 3.89 & -0.02 & 10.90 \\ \cdashline{2-8}
 & \mdistilbert & 74.98 & 57.41 & 0.05 & 2.45 & 0.31 & 8.92 \\ 
 & \mminilm & 74.42 & 69.17 & 0.00 & 3.17 & 0.16 & 9.50 \\ \midrule
\massive & \mbert & 55.65 & 50.41 & 0.43 & 2.37 & 5.76 & 21.63 \\
 & \xlmr & 75.10 & 65.55 & 0.42 & 2.80 & 6.75 & 26.42 \\
 & \rembert & 83.83 & 73.29 & 0.40 & 3.00 & 6.17 & 27.51 \\
 & \mdeberta & 84.05 & 73.83 & 0.40 & 3.04 & 6.43 & 28.03 \\ \cdashline{2-8}
 & \mdistilbert & 47.20 & 42.68 & 0.32 & 2.01 & 4.38 & 19.10 \\
 & \mminilm & 57.91 & 54.72 & 0.29 & 2.44 & 5.11 & 23.13 \\ \midrule
\mtop & \mbert & 69.17 & 67.59 & 0.60 & 2.57 & 9.30 & 26.89 \\
 & \xlmr & 88.40 & 77.84 & 0.59 & 2.95 & 8.70 & 30.57 \\
 & \rembert & 91.73 & 79.69 & 0.58 & 3.05 & 8.82 & 31.35 \\
 & \mdeberta & 91.24 & 79.78 & 0.59 & 2.97 & 8.19 & 30.70 \\ \cdashline{2-8}
 & \mdistilbert & 60.21 & 59.73 & 0.45 & 2.46 & 7.22 & 25.13 \\
 & \mminilm & 76.14 & 70.60 & 0.52 & 2.77 & 7.13 & 29.12 \\

\midrule
Mean &  & 77.24 & 70.02 & 0.33 & 2.83 & 4.70 & 21.46 \\
\bottomrule
\end{tabular}%
} 

\caption{The overall results for intent recognition and slot filling on test sets in German and the gap in performance before and after dialect perturbations are applied. Setup \textit{(ii)}: English train set is used for training, German development set is used for model selection. Intact (left): performance on intact test sets. Individual perturbations (middle): 18 individual perturbations are applied and average performance gap is computed across them. All perturbations (right): all perturbations applied simultaneously. $\Delta$ denotes the difference between performance on intact and perturbed data. Performance on intact data consistently surpasses that on perturbed data, leading to positive $\Delta$ values. The results are averaged across five runs with varying random initialization.}
\label{tab:ende_results}
\end{table*}

\newpage
\clearpage

\subsection{Performance on perturbed German test sets in fine-tuning setup}

\begin{table*}[h]
\resizebox{0.95\textwidth}{!}{ %
\begin{tabular}{llcccccc}
\toprule
& &  \multicolumn{2}{c}{Intact} &  \multicolumn{2}{c}{Individual Perturbations} &  \multicolumn{2}{c}{All Perturbations} \\  
&   &   Intent Acc  &  Slot F$_{1}$ &  $\Delta$ Intent Acc  & $\Delta$ Slot F$_{1}$ & $\Delta$  Intent Acc & $\Delta$   Slot F$_{1}$  \\  
\cmidrule(lr){3-4}  \cmidrule(lr){5-6}  \cmidrule(lr){7-8}
\atis & \mbert & 79.17 & 92.66 & 0.01 & 5.05 & 0.33 & 4.24 \\
 & \xlmr & 79.17 & 92.56 & 0.01 & 4.97 & 0.33 & 3.09 \\
 & \rembert & 79.22 & 92.33 & 0.01 & 4.97 & 0.33 & 3.52 \\
 & \mdeberta & 79.46 & 92.55 & 0.01 & 4.99 & 0.33 & 3.08 \\ \cdashline{2-8}
 & \mdistilbert & 79.28 & 92.12 & 0.01 & 5.09 & 0.33 & 5.44 \\
 & \mminilm & 75.70 & 89.00 & 0.00 & 4.71 & 0.11 & 8.27 \\ \midrule
\massive & \mbert & 84.79 & 77.38 & 0.37 & 3.01 & 6.22 & 6.98 \\
 & \xlmr & 86.73 & 78.83 & 0.34 & 3.06 & 5.78 & 6.95 \\
 & \rembert & 87.22 & 80.05 & 0.33 & 3.17 & 5.71 & 6.98 \\ 
 & \mdeberta & 87.19 & 79.62 & 0.35 & 3.12 & 5.95 & 6.96 \\ \cdashline{2-8}
 & \mdistilbert & 83.25 & 76.60 & 0.35 & 2.98 & 5.68 & 7.13 \\ 
 & \mminilm & 80.09 & 76.43 & 0.33 & 2.99 & 5.48 & 6.99 \\ \midrule
\mtop & \mbert & 94.64 & 83.74 & 0.49 & 3.04 & 7.46 & 7.36 \\
 & \xlmr & 95.71 & 84.37 & 0.48 & 3.07 & 7.43 & 6.95 \\
 & \rembert & 95.98 & 84.19 & 0.49 & 3.10 & 7.86 & 7.27 \\
 & \mdeberta & 95.62 & 84.55 & 0.48 & 3.07 & 7.86 & 6.96 \\ \cdashline{2-8}
 & \mdistilbert & 93.94 & 82.05 & 0.50 & 3.02 & 7.63 & 7.35 \\
 & \mminilm & 89.78 & 82.77 & 0.55 & 3.03 & 8.48 & 7.36 \\ 

\midrule
Mean &  & 85.94 & 84.54 & 0.28 & 3.69 & 4.63 & 6.27 \\ 
\bottomrule
\end{tabular}%
} 

\caption{The overall results for intent recognition and slot filling on test sets in German and the gap in performance before and after dialect perturbations are applied. Setup \textit{(iii)}: German train set is used for training; German development set is used for model selection. Intact (left): performance on intact test sets. Individual perturbations (middle):  18 individual perturbations are applied and average performance gap is computed across them. All perturbations (right): all perturbations applied simultaneously. $\Delta$ denotes the difference between performance on intact and perturbed data. Performance on intact data consistently surpasses that on perturbed data, leading to positive $\Delta$ values. The results are averaged across five runs with varying random initialization.}
\label{tab:dede_results}
\end{table*}

\newpage
\clearpage

\section{Success rate} \label{sec:success_rate}

\subsection{Success rate of English perturbations}
\begin{figure}[htp!]

\begin{tabularx}{\linewidth}{CC}
    \begin{subfigure}{\linewidth}
        \includegraphics[width=1.\linewidth]{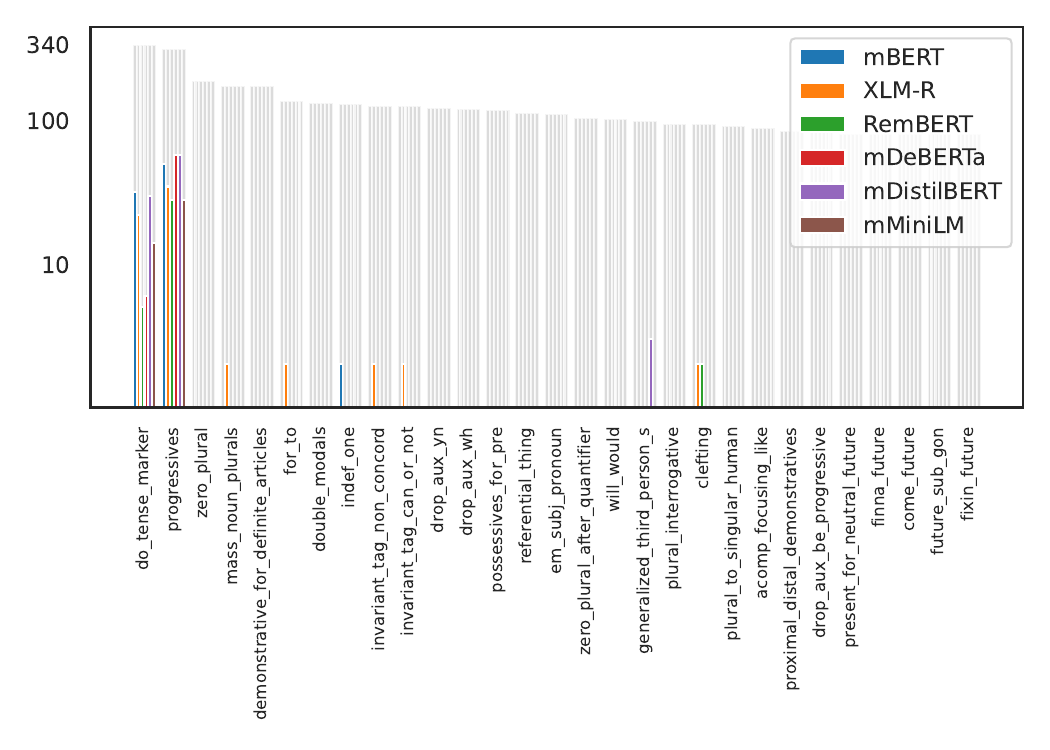}
    \caption{\xsid}
    \end{subfigure}

    \begin{subfigure}{\linewidth}
        \includegraphics[width=1.\linewidth]{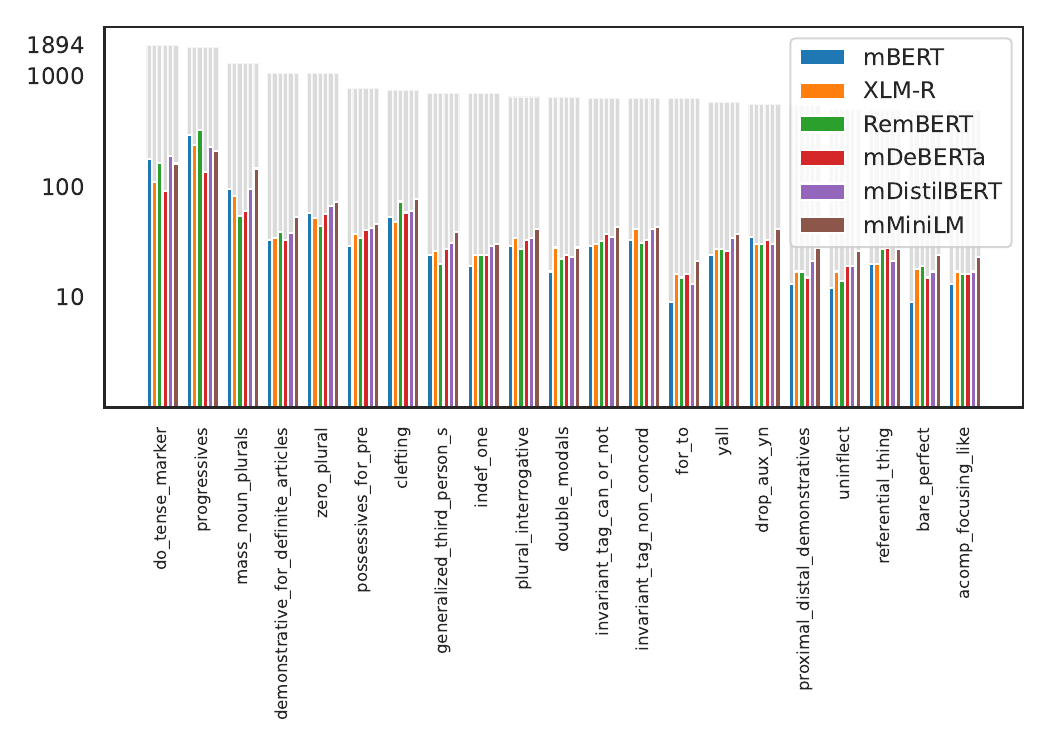}
    \caption{\massive}
    \end{subfigure}

& 

    \begin{subfigure}{\linewidth}
        \includegraphics[width=1.\linewidth]{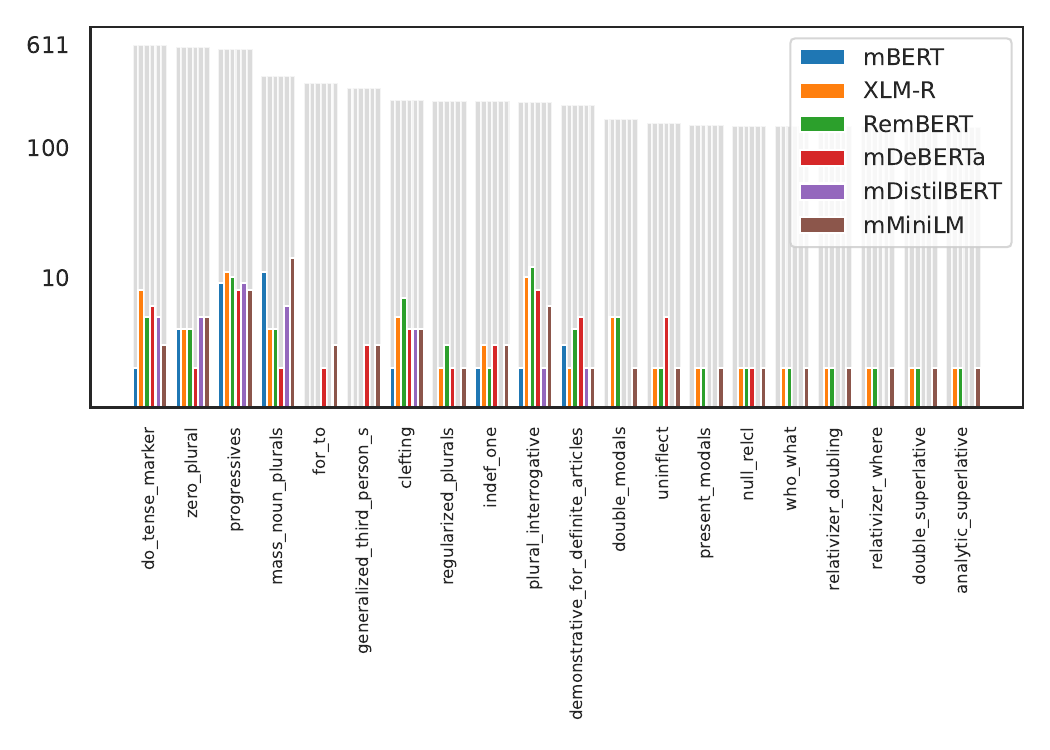}
    \caption{\atis}
    \end{subfigure}

    \begin{subfigure}{\linewidth}
        \includegraphics[width=1.\linewidth]{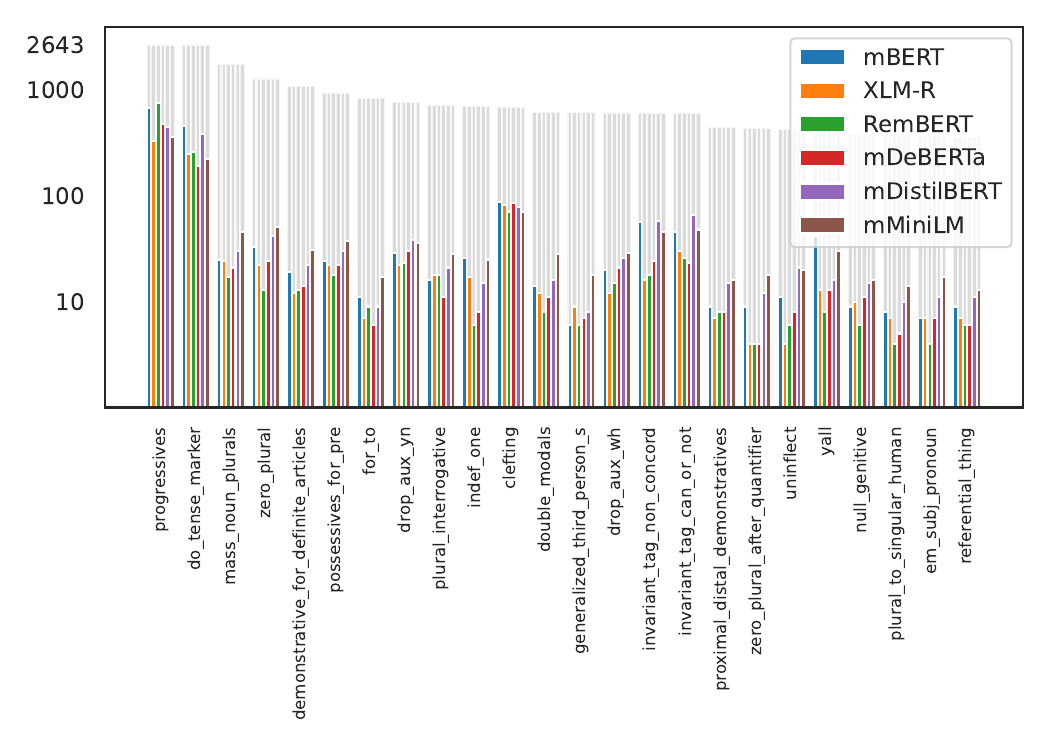}
    \caption{\mtop}
    \end{subfigure}

\end{tabularx}
\caption{The success rates in intent prediction on the perturbed English tests sets  with respect to individual perturbations. The grey bars represent the perturbation frequency (i.e., the count of altered sentences), while the colored bars indicate the success rate (i.e., the number of misclassified sentences after applying the perturbation). A logarithmic scale is utilized for improved clarity.}
\label{fig:en_success_rate}
\end{figure}

\newpage
\clearpage

\subsection{Success rate of German perturbations}
\begin{figure}[htp!]

\begin{tabularx}{\linewidth}{CC}
    \begin{subfigure}{\linewidth}
        \includegraphics[width=1.\linewidth]{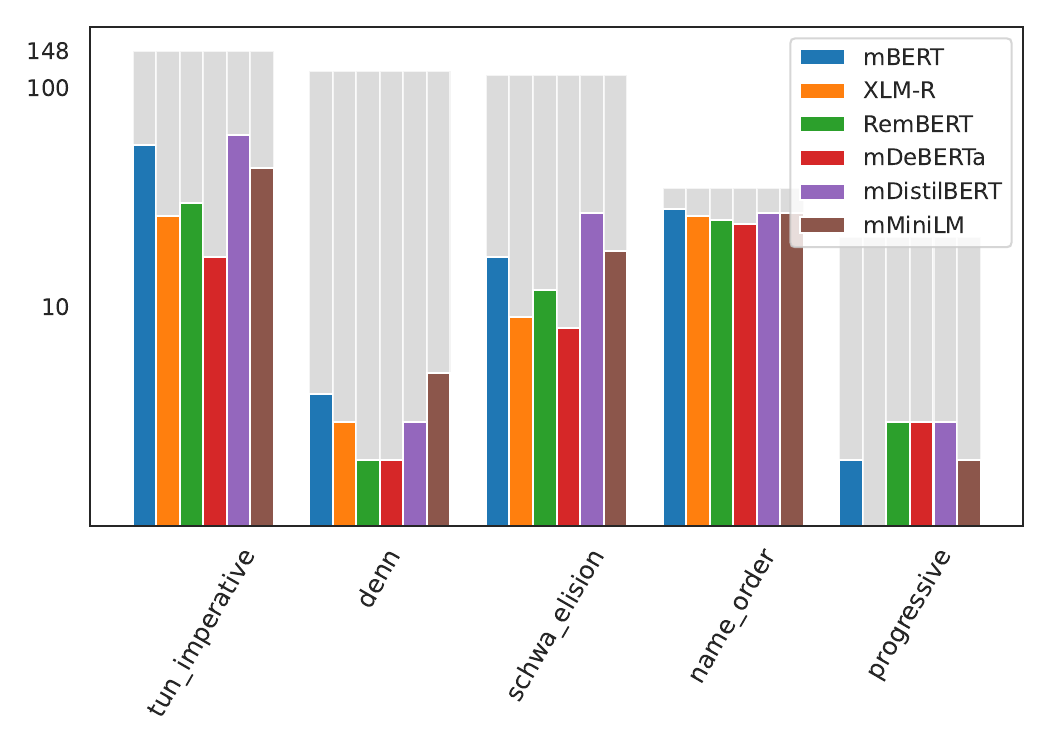}
    \caption{\xsid}
    \end{subfigure}

    \begin{subfigure}{\linewidth}
        \includegraphics[width=1.\linewidth]{plots/de.MASSIVE_success_rate.pdf}
    \caption{\massive}
    \end{subfigure}

& 

    \begin{subfigure}{\linewidth}
        \includegraphics[width=1.\linewidth]{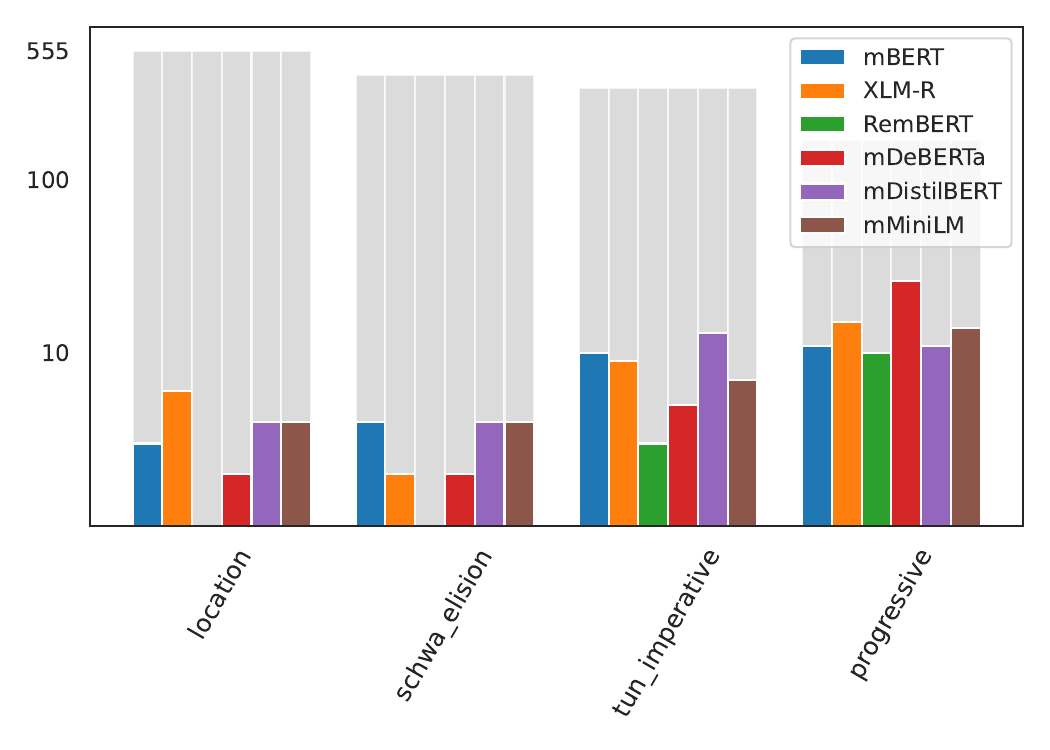}
    \caption{\atis}
    \end{subfigure}

    \begin{subfigure}{\linewidth}
        \includegraphics[width=1.\linewidth]{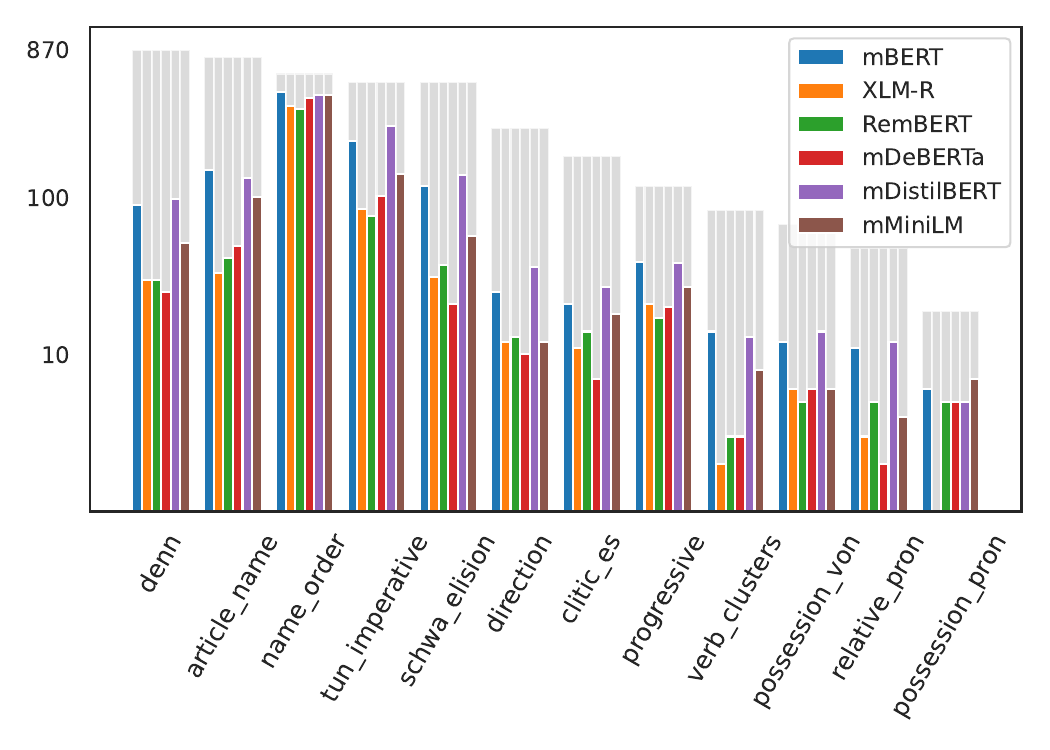}
    \caption{\mtop}
    \end{subfigure}

\end{tabularx}
\caption{The success rates in intent prediction on the perturbed German tests sets  with respect to individual perturbations. The grey bars represent the perturbation frequency (i.e., the count of altered sentences), while the colored bars indicate the success rate (i.e., the number of misclassified sentences after applying the perturbation). A logarithmic scale is utilized for improved clarity.}
\label{fig:de_success_rate}
\end{figure}

\newpage
\clearpage

\section{Evaluation of performance drop} \label{sec:f1_drop_cat}

\subsection{Performance drop in English} \label{sec:en_f1_drop_cat}
\begin{figure}[htp!]

\begin{tabularx}{\linewidth}{CC}
    \begin{subfigure}{\linewidth}
        \includegraphics[width=1.\linewidth]{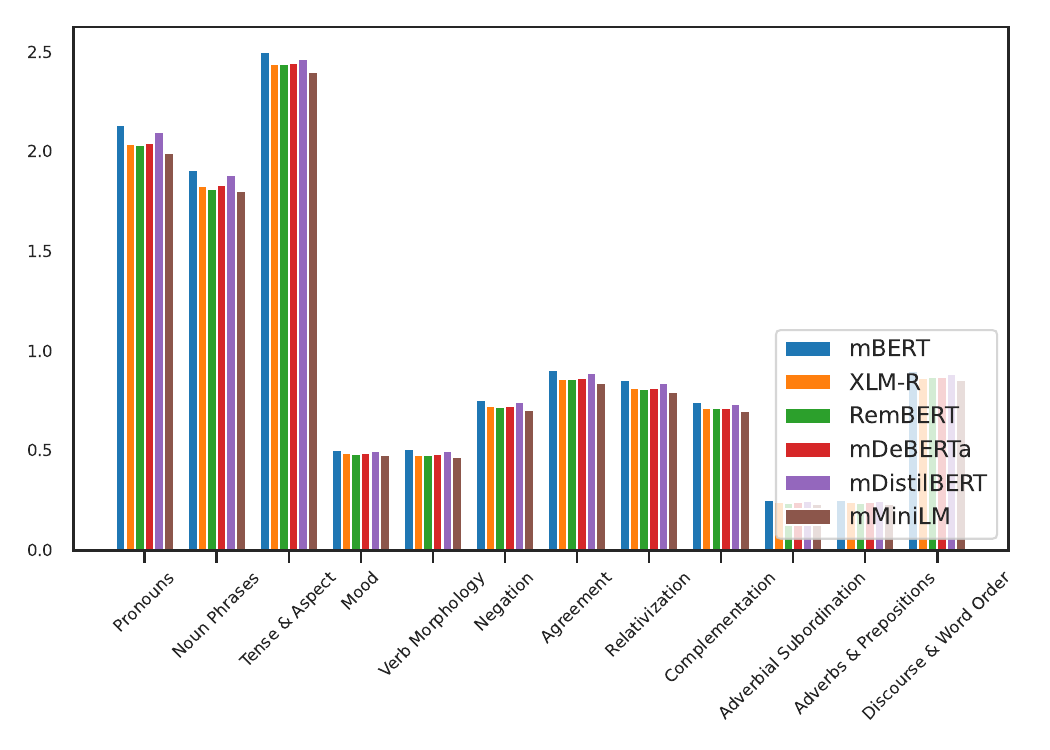}
    \caption{\xsid}
    \end{subfigure}

    \begin{subfigure}{\linewidth}
        \includegraphics[width=1.\linewidth]{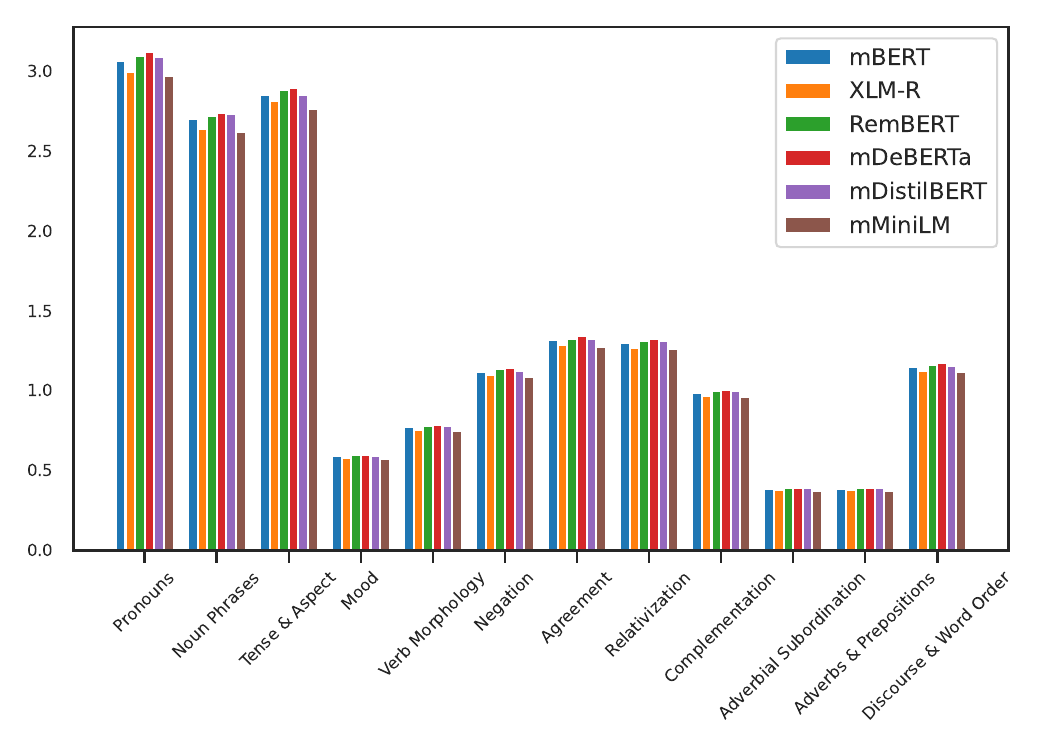}
    \caption{\massive}
    \end{subfigure}

& 

    \begin{subfigure}{\linewidth}
        \includegraphics[width=1.\linewidth]{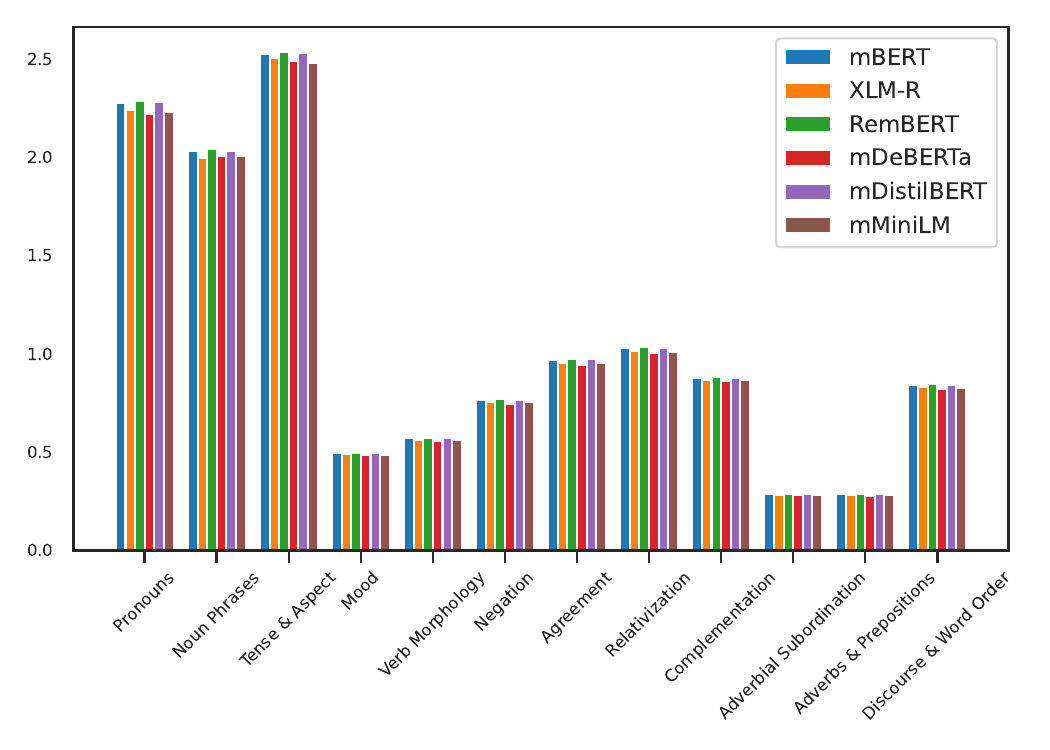}
    \caption{\atis}
    \end{subfigure}

    \begin{subfigure}{\linewidth}
        \includegraphics[width=1.\linewidth]{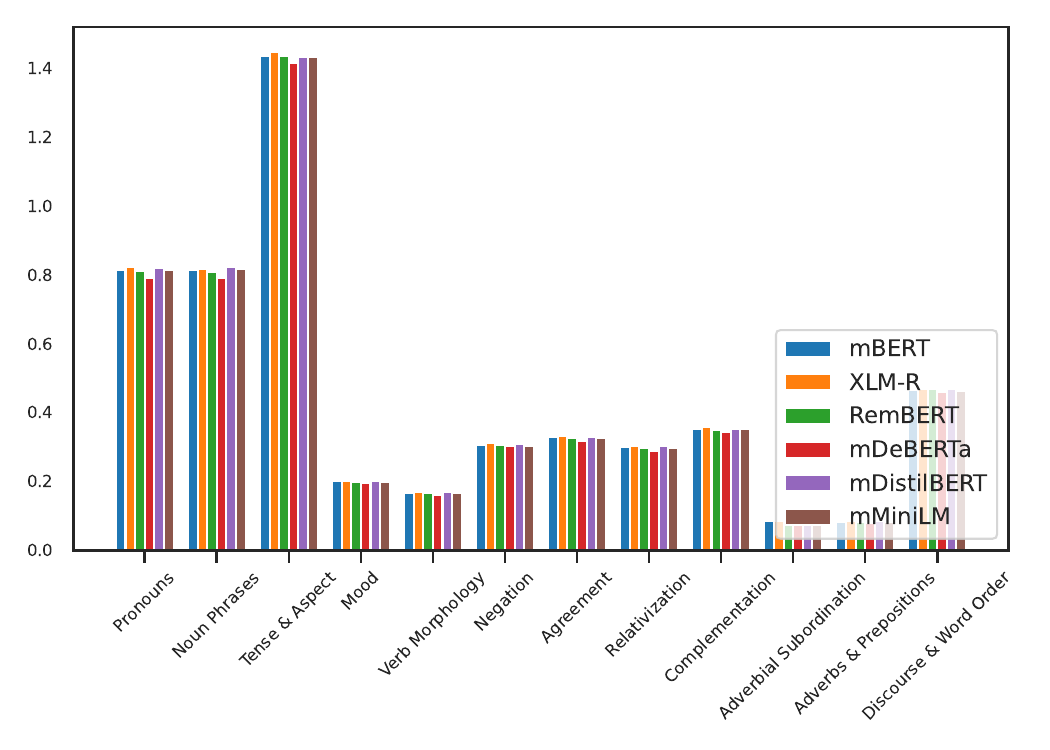}
    \caption{\mtop}
    \end{subfigure}

\end{tabularx}
\caption{The $\Delta$ \textit{F$_1$}  with respect to perturbation category in perturbed English test sets.}
\label{fig:en_f1_drop_cat}
\end{figure}

\newpage
\clearpage

\subsection{Performance drop in German}
\begin{figure}[htp!]

\begin{tabularx}{\linewidth}{CC}
    \begin{subfigure}{\linewidth}
        \includegraphics[width=1.\linewidth]{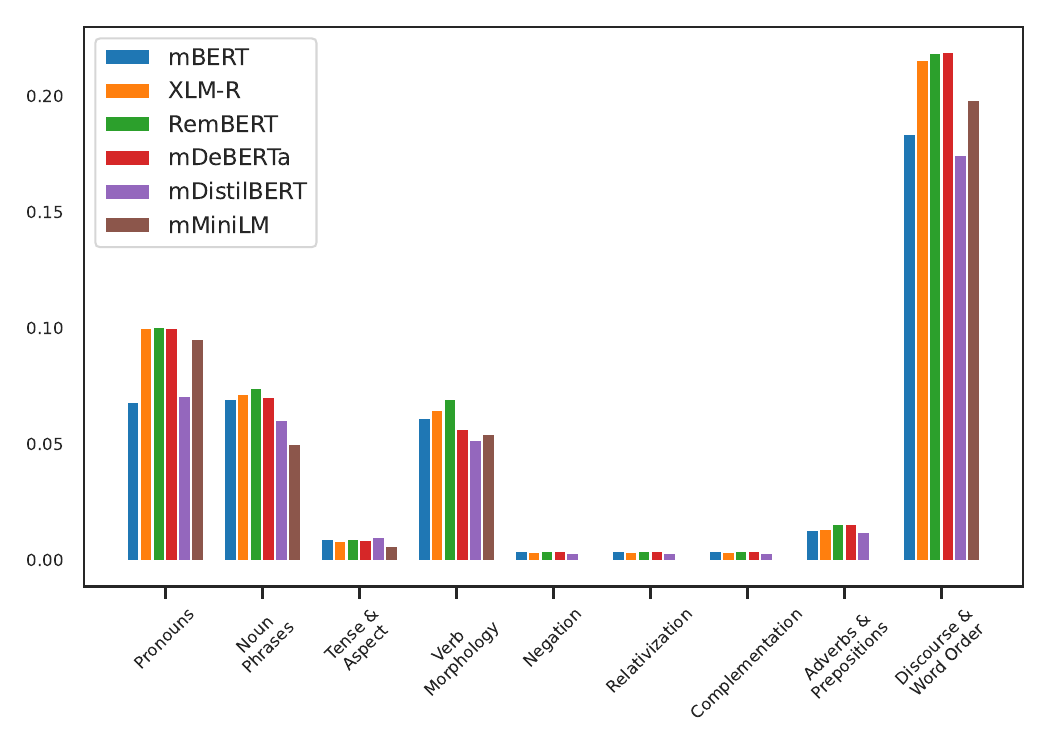}
    \caption{\xsid}
    \end{subfigure}

    \begin{subfigure}{\linewidth}
        \includegraphics[width=1.\linewidth]{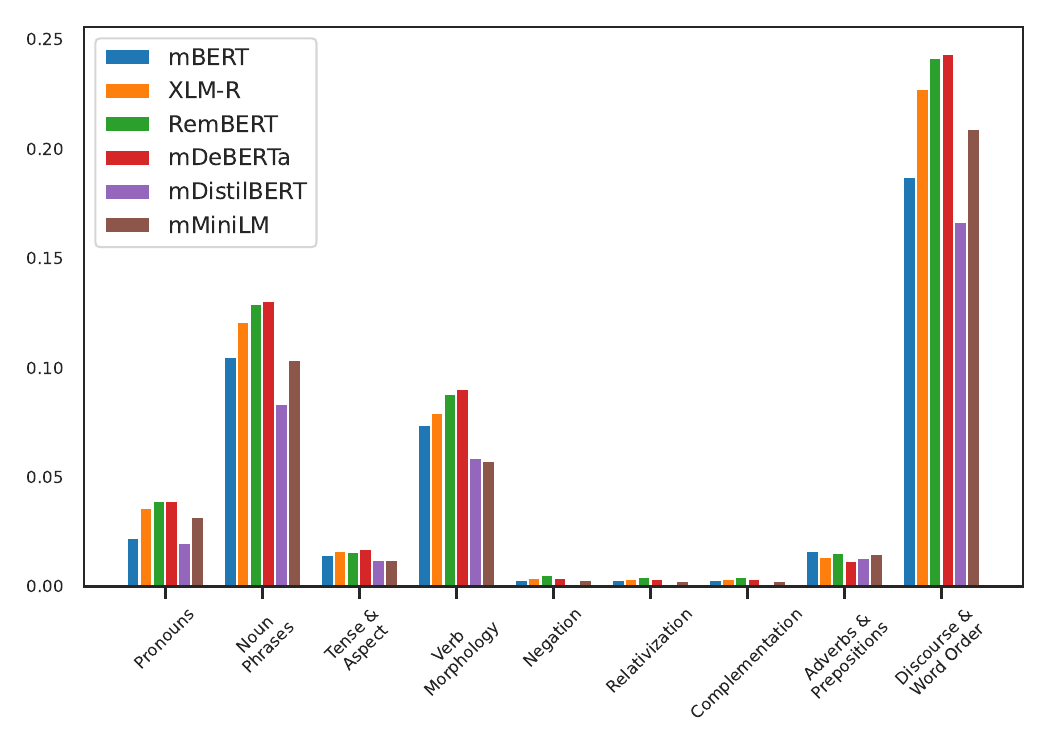}
    \caption{\massive}
    \end{subfigure}

& 

    \begin{subfigure}{\linewidth}
        \includegraphics[width=1.\linewidth]{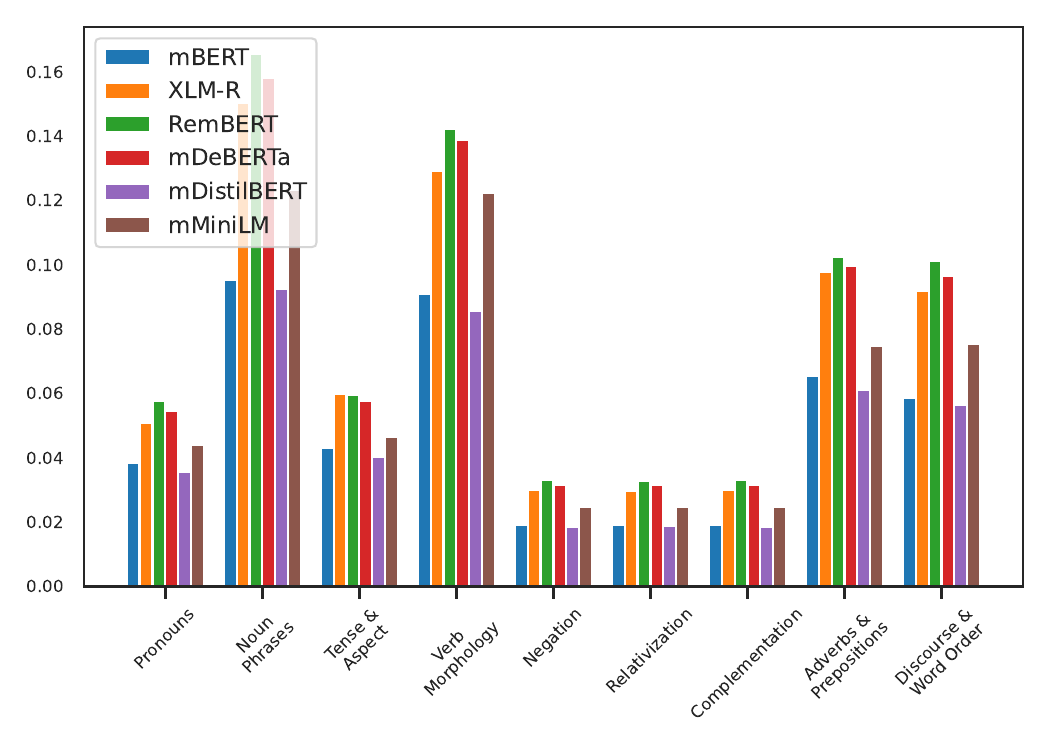}
    \caption{\atis}
    \end{subfigure}

    \begin{subfigure}{\linewidth}
        \includegraphics[width=1.\linewidth]{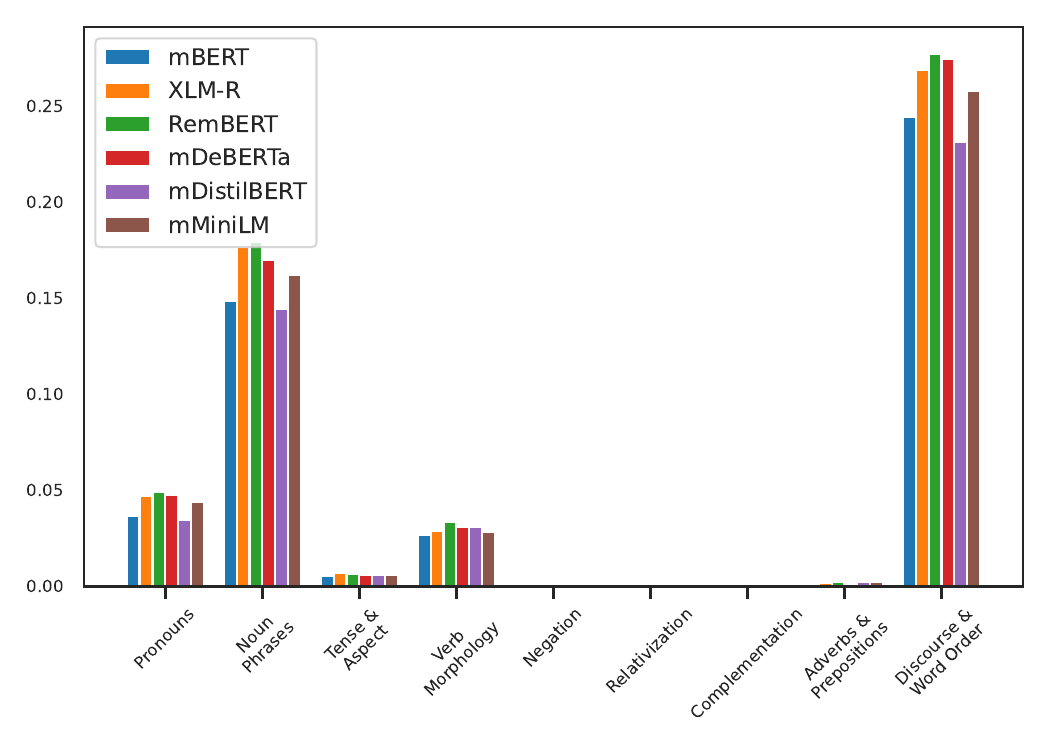}
    \caption{\mtop}
    \end{subfigure}

\end{tabularx}
\caption{The $\Delta$ \textit{F$_1$}  with respect to perturbation category in perturbed German test sets.}
\label{fig:de_f1_drop_cat}
\end{figure}

\newpage
\clearpage

\section{Human evaluation guidelines} \label{sec:instruction}
\subsection*{Sentence Pair Assessment}

You'll be given a pair of sentences. One is in standard German, and the other is a re-write in dialect or colloquial German.  A is for German sentences,  B is for dialect re-writes. 

Your job is to rate the naturalness and fluency of the re-write on a scale of one to five. Does the re-write sound like something you could say? A score of one indicates that the re-write sounds unnatural, while a score of five means that the re-write is fluent and completely acceptable. Trust your gut feeling and don't overthink it. If you're unsure about the score, choose the ``idk'' option (I don't know). Feel free to add comments if necessary.

\paragraph{Example}
\begin{itemize}
    \item[A] Ich muss Papa jetzt anrufen .	
    \item[B] Ich muss den Papa jetzt anrufen .
\end{itemize}

\begin{tabular}[t]{|*{5}{c|}}%
\hline
1 - bad & 2 & 3 & 4 & 5 - great  \\ \hline
\multicolumn{5}{|l|}{Comments (free form): } \\
\hline
\end{tabular}

The information from your evaluation will only be used for research. 

Thank you for your time and effort!

\end{document}